%% file: paper.tex
\documentclass{arxivpaper}

\usepackage{xspace}
\usepackage{xcolor}
\usepackage[normalem]{ulem}
\usepackage{adjustbox}
\usepackage{enumitem}
\usepackage{listings}
\usepackage{array}

\makeatletter
\DeclareRobustCommand\onedot{\futurelet\@let@token\@onedot}
\def\@onedot{\ifx\@let@token.\else.\null\fi\xspace}

\makeatother

\newtcolorbox{graylist}{
  enhanced,
  colback=gray!5,
  colframe=gray!15,
  boxrule=0pt,
  arc=0mm,
  left=1mm, right=1mm, top=1mm, bottom=1mm,
  width=\dimexpr\linewidth-4mm\relax,
  center,
  breakable
}

\useunder{\uline}{\ul}{}

\paperlogo[1.5cm]{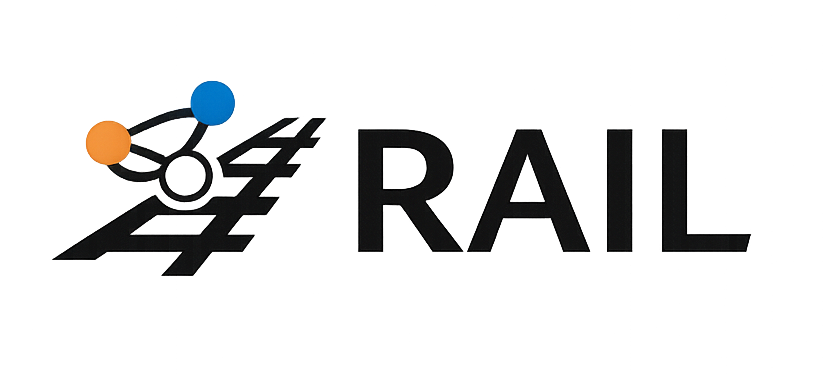}

\title{Are LLMs Positionally Consistent Ordinal Classifiers? A Systematic Evaluation}
\input{sec/0_abstract}

\author[1]{Yu Wang}
\author[2]{Zhe Zhou}
\author[3]{Menglin Liu}
\author[4]{Ge Shi\thanks{Corresponding author: geshi@ucdavis.edu}}

\affiliation[1]{Cornell University}
\affiliation[2]{University of Washington}
\affiliation[3]{The Chinese University of Hong Kong, Shenzhen}
\affiliation[4]{University of California, Davis}

\begin{document}

\maketitle

\input{sec/1_introduction}
\input{sec/2_related_work}
\input{sec/3_experimental_framework}
\input{sec/4_phase_0_prevalence_of_position_bias}
\input{sec/5_phase1}
\input{sec/6_phase2}
\input{sec/7_conclusion}

\bibliographystyle{assets/plainnat}
\bibliography{paper}

\appendix
\input{appendix/appendix}

\newpage

\end{document}

%% file: sec/0_abstract.tex
\abstract{
Large language models are increasingly used for ordinal classification, yet semantically equivalent changes to prompt organization can alter their predictions. We conduct systematic experiments to characterize positional bias from label order, demonstration order, and demonstration placement. First, we apply the three probes to ten frontier LLMs on a common ordinal-classification task; every model is sensitive to all three positional sources, showing that the problem is pervasive. Second, we vary eight prompt-, task-, and model-level factors across five datasets; accuracy and stability are often misaligned, and only lower scale cardinality consistently improves both. Third, we compare pointwise, pairwise, and listwise inference, alternative aggregation and debiasing methods, and joint configurations; the tested corrections do not provide a reliable remedy, while a comparison-based listwise formulation offers the best balance but transfers unevenly across models and bias sources. These findings show that positional robustness depends on the full system configuration rather than the model alone. Ordinal-classification systems should therefore be selected jointly for predictive performance and stability.
}

%% file: sec/1_introduction.tex
\section{Introduction}

Large language models are increasingly used as classifiers through in-context learning, but their predictions can change under prompt variations that should not alter the task. Prior work documents sensitivity to label or option identifiers, prompt formatting, and demonstration order~\citep{min2022rethinking,sclar2024quantifying,lu2022fantastically,kumar2021reordering,wei2024unveiling,yang2025option}. Position bias is one manifestation: predictions depend on option order and identifiers~\citep{pezeshkpour2024large,zheng2024large,wei2024unveiling,yang2025option}, the arrangement of demonstrations~\citep{lu2022fantastically,kumar2021reordering}, or their placement relative to the query~\citep{cobbina2025does}.

\begin{figure*}[t]
  \centering
  \includegraphics[width=0.98\textwidth]{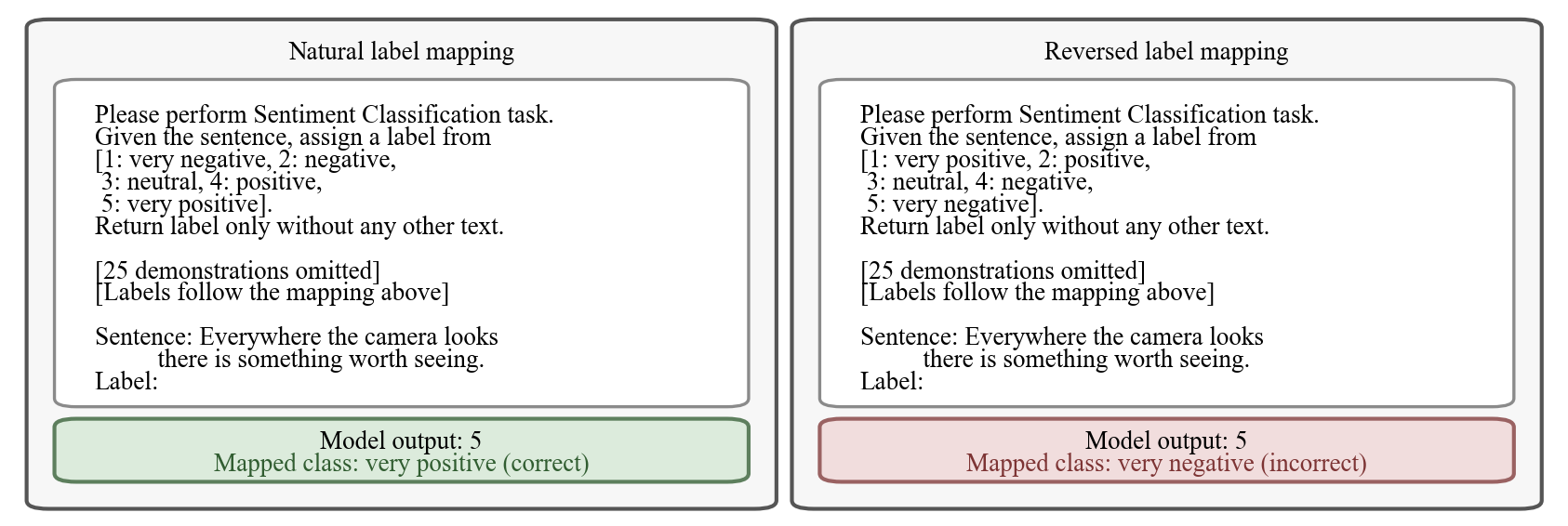}
  \vspace{-2pt}
  \caption{Illustrative label-order sensitivity in SST-5. The prompts differ only in the numeric mapping and remapped demonstration labels; demonstrations are omitted. Repeating output 5 changes the prediction from very positive to very negative.}
  \label{fig:position_bias_example}
  \vspace{-7pt}
\end{figure*}

To our knowledge, most position-bias evidence for classification comes from nominal multiple-choice tasks, where labels have no inherent order~\citep{pezeshkpour2024large,zheng2024large,wei2024unveiling,yang2025option}. Ordinal classification instead predicts along a ranked scale, as in sentiment intensity and review scoring~\citep{gutierrez2016ordinal}. Here, distant mistakes are more consequential than adjacent ones, motivating MAE alongside accuracy, while Spearman's correlation captures rank agreement~\citep{baccianella2009evaluation}. Positional instability can therefore distort individual labels, rankings, and scalar measurements~\citep{licht2025measuring}. Figure~\ref{fig:position_bias_example} illustrates the simplest case: repeating the same numeric output after reversing the label mapping changes its semantic meaning.

Position bias is well documented in pairwise and listwise LLM judges~\citep{wang2024llms,shi2025judging}. Rubric-based and scalar scoring additionally exhibit order-sensitive score selection, prompt-induced scoring shifts, and numerical-output artifacts~\citep{xu2026pointwise,li2026scoringbias,licht2025measuring}. The most directly comparable work in ordinal classification, LAMPO~\citep{qin2024lampo}, compares each test instance with individual demonstrations and swaps their order to reduce pairwise position bias. It does not systematically probe label-mapping order, demonstration ordering, or placement, broader prompt factors, competing pipelines, or transfer across models. We address this gap with three probes that isolate label order, demonstration order, and demonstration placement.

\noindent\textbf{Pervasiveness.} Across ten frontier LLMs from seven providers, all 30 model--probe combinations show significant instability under greedy decoding, with prediction flip rates (PFRs) from 8.5\% to 61.5\%. Under stochastic decoding, positional bias remains distinguishable from repeated-run noise for six models.

\noindent\textbf{Factors.} Controlled ablations across eight prompt- and model-level factor families show that accuracy and stability are often decoupled. Natural-language labels and minimal instructions raise label-order PFR by 59.5 and 48.8 points while changing accuracy by only $+1.3$ and $-3.7$ points; among the tested factors, reducing scale cardinality is the only intervention that consistently improves both performance and stability.

\noindent\textbf{Mitigation and transfer.} PriDe~\citep{zheng2024large} and Contextual Calibration~\citep{zhao2021calibrate} do not reduce label-order bias, while pairwise inference~\citep{qin2024lampo} and direct listwise inference adapted from learning-to-rank formulations~\citep{liu2009learning,sun2023chatgpt} sacrifice performance. Our listwise-compare formulation reduces label-order PFR by 18.1 points with little accuracy loss. Joint prompt configurations selected on Qwen3~\citep{yang2025qwen3} transfer unevenly and can trade one form of positional stability for another.

Together, these results show that ordinal-classification prompts and pipelines should be selected as multi-objective robustness configurations rather than by accuracy under a single prompt arrangement.

%% file: sec/2_related_work.tex
\section{Related Work}

In-context learning is sensitive to prompt formatting, label surface forms, and the selection and order of demonstrations; even replacing demonstration labels with incorrect ones can preserve substantial performance, showing that prompt-level regularities contribute alongside task semantics~\citep{zhao2021calibrate,sclar2024quantifying,lu2022fantastically,min2022rethinking}. These sensitivities motivate treating prompt design as an experimental factor rather than a fixed implementation detail.

Multiple-choice question (MCQ) studies identify systematic preferences for option positions, identifiers, and symbols, with effects varying across models~\citep{pezeshkpour2024large,zheng2024large,wei2024unveiling,yang2025option}. Related position and scoring biases occur in LLM judges~\citep{wang2024llms,shi2025judging,xu2026pointwise}. Ordinal classification remains comparatively underexplored: LAMPO decomposes prediction into pairwise comparisons between the test instance and individual demonstrations and swaps pair order to reduce comparison-position bias, but does not systematically probe label-mapping order, demonstration ordering, or placement~\citep{qin2024lampo}.

Proposed mitigations include contextual, domain-context, and batch calibration; position-prior correction; permutation aggregation or training; and distribution-level correction~\citep{zhao2021calibrate,fei2023mitigating,zhou2024batch,zheng2024large,tang2024found,liusie2024teacher,li2025calibraeval}. Most are evaluated within a particular bias source or task setting, whereas we jointly measure label order, demonstration order, and demonstration placement in ordinal ICL and compare prompt-level and pipeline-level interventions under a common framework.

%% file: sec/3_experimental_framework.tex
\section{Measuring Position Bias}

This section defines the task, datasets, positional probes, and evaluation metrics used throughout the study.

\subsection{Task and Datasets}

We study \(m\)-way ordinal classification through \(K\)-shot-per-class in-context learning, where \(m\) is the number of ordered classes and \(K\) is the number of labeled demonstrations sampled for each class. Each prompt therefore contains \(D=mK\) demonstrations, followed by one test input that the model assigns to one of the \(m\) classes. We use five datasets spanning \(m\in\{2,3,5\}\) and two task domains, summarized in Table~\ref{tab:datasets}. Unless the demonstration count is being ablated, we set \(K=5\), giving \(D=10\), \(15\), or \(25\) demonstrations for the two-, three-, and five-way datasets, respectively. Phase~0 fixes SST-5 (\(m=5,D=25\)) to compare ten frontier models under a common task, whereas Phases~1 and 2 aggregate results over all applicable datasets and three demonstration seeds. Each dataset uses a fixed stratified test set of 200 instances. 

\begin{table*}[!tbp]
\centering
\small
\setlength{\tabcolsep}{3pt}
\begin{tabular}{@{}>{\raggedright\arraybackslash}p{0.293\textwidth}cc>{\raggedright\arraybackslash}p{0.144\textwidth}>{\raggedright\arraybackslash}p{0.42\textwidth}@{}}
\toprule
Dataset & $m$ & $D$ & Domain & Labels (ordinal order) \\
\midrule
SST-5 (Socher et al.,~\citeyear{socher2013recursive}) & 5 & 25 & Movie reviews & very negative, negative, neutral, positive, very positive \\
Yelp-5 (Zhang et al.,~\citeyear{zhang2015character}) & 5 & 25 & Business reviews & very negative, negative, neutral, positive, very positive \\
Twitter (Rosenthal et al.,~\citeyear{rosenthal2017semeval}) & 3 & 15 & Social media & negative, neutral, positive \\
Hate (Basile et al.,~\citeyear{basile2019semeval}) & 2 & 10 & Social media & non-hate, hate \\
Offensive (Zampieri et al.,~\citeyear{zampieri2019semeval}) & 2 & 10 & Social media & non-offensive, offensive \\
\bottomrule
\end{tabular}
\vspace{-2pt}
\caption{Overview of datasets. Access and licensing details appear in Supplementary Section~\ref{app:dataset_access}.}
\label{tab:datasets}
\vspace{-7pt}
\end{table*}

\subsection{Position Bias: Sources and Measurement}

Position bias occurs when changing the positional arrangement of prompt components causes a model's prediction to change, even though the semantic content remains the same. We measure it through three independent probes, each targeting a different source of positional sensitivity. Each probe varies one positional dimension while holding the others at their base values.

Probe 1 measures model sensitivity to the mapping between numeric labels and semantic classes. We run the model twice, once with each mapping. Taking SST-5 as an example, in the natural order run, labels are assigned as 1: very negative, 2: negative, 3: neutral, 4: positive, 5: very positive. In the reversed run, the mapping is flipped: 1: very positive, 2: positive, 3: neutral, 4: negative, 5: very negative. Both the label-space description in the prompt and the demonstration labels are reversed together. Predictions from the reversed run are mapped back to the original class space before comparison.

Probe 2 targets the ordering of few-shot demonstrations within the prompt. We construct base, class-ascending, class-descending, and fixed-random orderings. Taking SST-5 as an example with 25 demonstrations (5 per class), the class-ascending order groups all class-1 demonstrations first, followed by class-2, and so on up to class-5. The class-descending order reverses this arrangement, while the random order shuffles the same 25 demonstrations with a fixed seed.

Probe 3 targets where the demonstration block is placed relative to the test input. We run the model under three placements. In the base placement, all demonstrations appear before the test input. In the after placement, all demonstrations appear after the test input. In the split placement, the first half of the demonstrations appear before the test input and the second half after.

For each experimental condition, we apply all three probes simultaneously. In some conditions, certain probes may not be applicable or require modified interpretation; we address these cases when reporting the corresponding results. Implementation details are provided in Supplementary Section~\ref{app:probe_config}.

\subsection{Performance and Bias Metrics}

\noindent\textbf{Performance metrics.} We report exact-match accuracy (Acc.$\uparrow$) and macro-averaged F1 (F1$\uparrow$). To capture ordinal performance, we also report Spearman's $\rho\uparrow$, which measures rank agreement between predicted and gold labels, and MAE$\downarrow$, which measures their average class distance. These metrics reveal ranking errors and distinguish distant mistakes from near-misses, aspects that are largely absent from position-bias evaluations centered on nominal multiple-choice tasks. We also report parse failure rate, the proportion of outputs that cannot be mapped to a valid label.

\noindent\textbf{Position-sensitivity metrics.} Following prior position-consistency and flip-rate formulations~\citep{shi2025judging,labruna2025positional,li2026scoringbias}, we report Prediction Flip Rate (PFR), the proportion of instances whose semantic prediction changes across positional conditions, whether or not the change improves accuracy. For a two-condition contrast, PFR is the complement of Position Consistency; our aggregate extension marks an instance as flipped when any condition changes its prediction. Lower values are better. P1$\downarrow$ compares natural and reversed label mappings. P2a$\downarrow$ and P2b$\downarrow$ compare random demonstration order with class-ascending and class-descending order, while P3a$\downarrow$ and P3b$\downarrow$ compare the before-query placement with the after and split placements. Phase~0 reports the aggregate extension for P2 and P3, whereas the factor analyses report pairwise contrasts.

\subsection{Statistical Tests}

In Phase 0, we use three statistical tests. The Wilson confidence interval provides reliable coverage for proportions even at extreme values or small sample sizes; we use it to test whether PFR is significantly greater than zero. McNemar's test compares two paired binary outcomes on the same instances; we use it both to compare accuracy across the two conditions of Probe~1 and, in Phase~0b, to test whether total flip indicators exceed repeated-run noise indicators. Cochran's Q test extends McNemar's test to more than two related groups; we use it to test whether accuracy differs across the multiple conditions of Probes~2 and 3. All tests use a significance level of $\alpha = 0.05$. In Phases~1 and 2, each condition is run across three demonstration seeds, and we report the mean and standard deviation.

%% file: sec/4_phase_0_prevalence_of_position_bias.tex
\section{Phase 0: Prevalence of Position Bias}

Phase~0 evaluates ten frontier LLMs from seven providers on SST-5 using all three probes (Table~\ref{tab:phase0}). All models are accessed through OpenRouter; model and API details appear in Supplementary Section~\ref{app:phase0_api}. Phase~0a uses greedy decoding to isolate positional effects, while Phase~0b uses stochastic decoding to compare positional sensitivity with repeated-run noise.

\subsection{Phase 0a: Deterministic Probing}

Phase 0a uses greedy decoding. Because greedy decoding is deterministic, any prediction differences across probe conditions are attributable solely to the positional manipulation.

\newcommand{\phasezeroci}[1]{{\setlength{\fboxsep}{1pt}\colorbox{gray!12}{\strut #1}}}
\newcommand{\phasezerocell}[2]{\shortstack{#1\\[-1pt]\phasezeroci{#2}}}
\newcommand{\phasezerosig}{\textcolor{green!50!black}{$\checkmark$}}
\newcommand{\phasezerons}{\textcolor{red!70!black}{$\times$}}

\begin{table*}[t]
\centering
\small
\renewcommand{\arraystretch}{0.90}
\setlength{\tabcolsep}{2pt}
\begin{tabular}{@{}lccc@{\hspace{8pt}}cccc@{}}
\toprule
& \multicolumn{3}{c}{\textbf{(a) Greedy decoding}} & \multicolumn{4}{c}{\textbf{(b) Stochastic decoding}} \\
\cmidrule(lr){2-4}\cmidrule(lr){5-8}
Model & P1 & P2 & P3 & Noise\textsuperscript{a} & Total\textsuperscript{b} & McNemar $p$ & Sig. \\
\midrule
Claude Haiku 4.5 & \phasezerocell{0.180}{[0.133, 0.239]} & \phasezerocell{0.240}{[0.186, 0.304]} & \phasezerocell{0.225}{[0.173, 0.288]} & \phasezerocell{0.050}{[0.027, 0.090]} & \phasezerocell{0.165}{[0.120, 0.223]} & $< 0.001$ & \phasezerosig \\
Claude Sonnet 4.5 & \phasezerocell{0.105}{[0.070, 0.155]} & \phasezerocell{0.180}{[0.133, 0.239]} & \phasezerocell{0.615}{[0.546, 0.680]} & \phasezerocell{0.000}{[0.000, 0.019]} & \phasezerocell{0.105}{[0.070, 0.155]} & $< 0.001$ & \phasezerosig \\
GPT-5.4 & \phasezerocell{0.085}{[0.054, 0.132]} & \phasezerocell{0.180}{[0.133, 0.239]} & \phasezerocell{0.325}{[0.264, 0.393]} & \phasezerocell{0.050}{[0.027, 0.090]} & \phasezerocell{0.100}{[0.066, 0.149]} & 0.022 & \phasezerosig \\
Gemini 3.1 Flash-Lite & \phasezerocell{0.195}{[0.146, 0.255]} & \phasezerocell{0.245}{[0.191, 0.309]} & \phasezerocell{0.200}{[0.151, 0.261]} & \phasezerocell{0.060}{[0.035, 0.102]} & \phasezerocell{0.185}{[0.137, 0.245]} & $< 0.001$ & \phasezerosig \\
Gemini 3.1 Pro & \phasezerocell{0.140}{[0.099, 0.195]} & \phasezerocell{0.100}{[0.066, 0.149]} & \phasezerocell{0.135}{[0.095, 0.189]} & \phasezerocell{0.045}{[0.024, 0.083]} & \phasezerocell{0.130}{[0.090, 0.184]} & 0.001 & \phasezerosig \\
DeepSeek V4 Flash & \phasezerocell{0.165}{[0.120, 0.223]} & \phasezerocell{0.300}{[0.241, 0.367]} & \phasezerocell{0.275}{[0.218, 0.341]} & \phasezerocell{0.130}{[0.090, 0.184]} & \phasezerocell{0.145}{[0.103, 0.201]} & 0.338 & \phasezerons \\
DeepSeek V4 Pro & \phasezerocell{0.180}{[0.133, 0.239]} & \phasezerocell{0.305}{[0.245, 0.372]} & \phasezerocell{0.340}{[0.278, 0.408]} & \phasezerocell{0.160}{[0.116, 0.217]} & \phasezerocell{0.210}{[0.159, 0.272]} & 0.067 & \phasezerons \\
Llama 4 Maverick & \phasezerocell{0.195}{[0.146, 0.255]} & \phasezerocell{0.330}{[0.269, 0.398]} & \phasezerocell{0.394}{[0.329, 0.463]} & \phasezerocell{0.060}{[0.035, 0.102]} & \phasezerocell{0.185}{[0.137, 0.245]} & $< 0.001$ & \phasezerosig \\
Qwen3.7-Plus & \phasezerocell{0.170}{[0.124, 0.228]} & \phasezerocell{0.120}{[0.082, 0.172]} & \phasezerocell{0.175}{[0.129, 0.234]} & \phasezerocell{0.135}{[0.095, 0.189]} & \phasezerocell{0.155}{[0.111, 0.212]} & 0.298 & \phasezerons \\
Grok 4.3 & \phasezerocell{0.095}{[0.062, 0.144]} & \phasezerocell{0.290}{[0.232, 0.356]} & \phasezerocell{0.190}{[0.142, 0.250]} & \phasezerocell{0.075}{[0.046, 0.120]} & \phasezerocell{0.095}{[0.062, 0.144]} & 0.240 & \phasezerons \\
\bottomrule
\end{tabular}
\vspace{-2pt}
\caption{Position sensitivity under greedy and stochastic decoding. Read each P1--P3, Noise, and Total cell as a prediction flip rate (PFR) estimate on the unshaded line and its Wilson 95\% confidence interval on the gray line. In panel (a), P1, P2, and P3 denote label order, demonstration order, and demonstration placement. Panel (b) uses temperature 0.7: Noise\textsuperscript{a} compares two runs of the same prompt, while Total\textsuperscript{b} compares natural and reversed label orders. Green checks indicate significance under a one-sided McNemar test at $p<0.05$; red crosses indicate nonsignificance.}
\label{tab:phase0}
\vspace{-7pt}
\end{table*}

Panel (a) reports PFRs with Wilson 95\% confidence intervals. All 30 intervals exclude zero, confirming universal position bias under greedy decoding. Even the smallest PFR (0.085 for GPT-5.4 on Probe 1) means roughly one in twelve predictions flips solely because the label order is reversed.

Bias severity varies considerably across probes and models. Probe 3 (demonstration placement) shows the largest variation, with PFRs ranging from 0.135 (Gemini 3.1 Pro) to 0.615 (Claude Sonnet 4.5). Probe 2 (demonstration order) exhibits a similar pattern, with Llama 4 Maverick and both DeepSeek models exceeding 0.30, whereas Gemini 3.1 Pro and Qwen3.7-Plus remain below 0.15. Probe 1 (label order) is comparatively stable, ranging from 0.085 to 0.195. Claude Sonnet 4.5 is particularly sensitive to demonstration placement: its aggregate P3 PFR reaches 0.615, and the after-query condition reduces accuracy by 26 percentage points, the largest effect observed.

\subsection{Phase 0b: Stochastic Decoding}

Phase 0a establishes position bias under deterministic decoding, but deployed LLMs typically use stochastic decoding. To distinguish positional effects from sampling noise, we compute a noise PFR by running the same prompt twice under identical settings and compare it with the total PFR (natural vs.\ reversed label order) using a one-sided McNemar test on the paired instance-level flip indicators.

Six of the ten models show total PFR significantly exceeding noise PFR ($p<0.05$), indicating that position bias persists beyond sampling noise. The remaining four (DeepSeek V4 Flash, DeepSeek V4 Pro, Qwen3.7-Plus, and Grok 4.3) do not reach significance—not because they lack position bias, which Phase 0a already establishes, but because stochastic decoding introduces enough noise to obscure it.

Supplementary Table~\ref{tab:hidden_tokens} reports hidden-token counts. Although reasoning is disabled, most models generate hidden tokens beyond the visible one-character response. The four models without significant differences between total and noise PFR also produce the largest hidden-token counts (156--512 per instance), and three have the highest noise PFRs in the study (0.130--0.160). This association suggests that extensive hidden-token generation may raise the noise floor, although it does not establish causation because model-specific decoding behavior or other unobserved factors could drive both.

\subsection{Summary of Phase 0}
Phase 0a establishes that position bias is universal under decoding, and, for most models, remains significantly larger than sampling noise under stochastic decoding. Phase 0b confirms that the four non-significant models also exhibit the largest hidden-token counts and generally higher noise PFRs, although this association is not causal. Having established the prevalence of position bias, we next examine the factors that modulate its severity.

%% file: sec/5_phase1.tex
\section{Factor Ablations}

We use one-factor-at-a-time ablations to isolate how prompt- and model-level choices affect positional sensitivity. Each condition changes one level relative to the starred baseline while holding all other settings fixed. Results are averaged over the applicable datasets and three demonstration seeds; the baseline and factor implementations appear in Supplementary Sections~\ref{app:baseline} and~\ref{app:phase1_config}.

\begin{table*}[!t]
\centering
\small
\setlength{\tabcolsep}{3pt}
\begin{tabular}{@{}ll>{\raggedright\arraybackslash}p{0.73\textwidth}@{}}
\toprule
Phase & Factor & Levels \\
\midrule
1 & Label format & Letter, Numeric*, Natural language, Neutral ID \\
1 & Model & Qwen3-8B Base/Instruct*/Thinking, Qwen3-4B, Gemma3-12B, Ministral-3-8B, Gemma4-12B, Granite-4-8B \\
1 & Scale cardinality & 5-way*, 3-way, 2-way \\
1 & Demo count & 0, 1, 3, 5* per class \\
1 & Prompt clarity & Minimal, Explicit*, Ordinal-explicit \\
1 & Separator & Colon+space*, space, tab \\
1 & Connector & Newline*, space, newline+tab \\
1 & Instruction mood & Imperative*, interrogative, indicative \\
\midrule
2 & Inference paradigm & Pointwise*, pairwise, listwise-compare, listwise \\
2 & Debiasing strategy & None*, Label averaging, Demo averaging, PriDe, Contextual Calibration \\
2 & Aggregation method & Copeland, majority vote, weighted sum*, mean score, Bradley--Terry \\
2 & Combined Phase 1 factors & Accuracy-optimal, P1-optimal, P2a-optimal, P2b-optimal \\
\bottomrule
\end{tabular}
\vspace{-2pt}
\caption{Factor families and configurations evaluated in this study. An asterisk (*) marks the baseline level for each factor.}
\label{tab:factor_taxonomy}
\vspace{-7pt}
\end{table*}

Table~\ref{tab:factor_taxonomy} summarizes the design. Phase~1 varies eight independently controlled factors; Phase~2 evaluates broader interventions that may change multiple pipeline components. Temperature, sampling, and label balance remain fixed, while context length changes only with demonstration count.

\subsection{Phase 1: Prompt- and Model-Level Factors}
\label{sec:phase1}

\input{sec/5_phase1_table}

Table~\ref{tab:phase1_residuals} reports mean percentage-point changes from the baseline. Label format most clearly separates accuracy from stability. Natural-language labels improve accuracy by only 1.3 points but increase label-order PFR by 59.5 points. Neutral identifiers instead improve accuracy by 6.5 points and reduce demonstration-order and placement PFRs by 7.3--13.0 points, while letter labels remain close to the numeric baseline.

Model effects are similarly mixed. Gemma4-12B improves accuracy and F1 by 13.4 and 15.5 points and reduces one placement PFR by 36.5 points; Gemma3-12B reduces all five PFRs. In contrast, Qwen3-8B Thinking improves accuracy by 3.0 points but raises label-order PFR by 9.4 points, while the base model improves accuracy by 7.2 points but raises placement PFRs by 20.1 and 26.2 points. Thus, stronger performance does not guarantee positional stability.

Scale cardinality produces the most consistent effect. Reducing the task from five classes to three or two raises accuracy by 25.0 and 43.9 points, respectively, while reducing every PFR by 16.3--48.9 points. Simpler label spaces are therefore both easier and more stable.

Demonstration count shows a different trade-off. Zero- and one-shot prompts improve accuracy by 7.1 and 4.4 points but raise label-order PFR by 6.9 and 3.4 points. When demonstrations are present, fewer examples generally reduce order and placement PFRs; one-shot reductions range from 6.2 to 20.2 points. These probes are undefined for zero-shot prompts.

Prompt clarity is asymmetric. Explicitly stating that labels are ordered changes accuracy by only +2.8 points and label-order PFR by +5.0 points, whereas removing the label-space description changes them by $-3.7$ and +48.8 points and also worsens $\rho$ and MAE. Essential task information matters more than additional explanation.

Low-level formatting primarily affects the second demonstration-order contrast. Replacing the separator with a space or tab lowers accuracy by 4.4--5.5 points and raises P2b by 11.2--15.6 points; a space connector has a similar effect, and an interrogative instruction raises P2b by 20.5 points. Semantically equivalent formatting can therefore substantially alter stability.

Overall, scale cardinality improves performance and stability together, while most other factors trade one objective or probe against another. Prompt selection should therefore evaluate accuracy and all positional probes jointly rather than optimize a single arrangement.

%% file: sec/5_phase1_table.tex
\definecolor{phasePositive}{HTML}{C65F5F}
\definecolor{phaseNegative}{HTML}{4C78A8}
\newcommand{\phasecell}[5]{\colorbox{#1!#2}{\textcolor{#3}{\makebox[3.65em][c]{$#4$}}}}
\newcommand{\phasena}{\colorbox{gray!20}{\makebox[3.65em][c]{--}}}

\begin{table*}[!t]
\centering
\small
\setlength{\tabcolsep}{0.8pt}
\renewcommand{\arraystretch}{1.0}
\begin{tabular}{l@{\hspace{4pt}}lccccccccc}
\toprule
& & \multicolumn{4}{c}{Performance residuals} & \multicolumn{5}{c}{Position-sensitivity residuals} \\
Factor & Level & Acc.$\uparrow$ & F1$\uparrow$ & $\rho\uparrow$ & MAE$\downarrow$ & P1$\downarrow$ & P2a$\downarrow$ & P2b$\downarrow$ & P3a$\downarrow$ & P3b$\downarrow$ \\
\midrule
Label format & Letter & \phasecell{phasePositive}{11}{black}{+2.20}{2.72} & \phasecell{phasePositive}{11}{black}{+2.71}{4.08} & \phasecell{phasePositive}{12}{black}{+3.06}{3.86} & \phasecell{phaseNegative}{10}{black}{-1.53}{3.72} & \phasecell{phasePositive}{12}{black}{+3.27}{5.15} & \phasecell{phasePositive}{9}{black}{+1.18}{7.38} & \phasecell{phasePositive}{10}{black}{+1.87}{6.52} & \phasecell{phasePositive}{13}{black}{+4.04}{15.33} & \phasecell{phasePositive}{10}{black}{+1.27}{5.32} \\
 & Natural language & \phasecell{phasePositive}{10}{black}{+1.27}{5.11} & \phasecell{phasePositive}{8}{black}{+0.35}{6.53} & \phasecell{phasePositive}{13}{black}{+4.01}{3.06} & \phasecell{phaseNegative}{10}{black}{-1.87}{5.93} & \phasecell{phasePositive}{80}{white}{+59.54}{27.81} & \phasecell{phaseNegative}{24}{black}{-13.18}{14.10} & \phasecell{phaseNegative}{10}{black}{-1.96}{4.70} & \phasecell{phasePositive}{32}{black}{+19.51}{24.07} & \phasecell{phasePositive}{13}{black}{+4.44}{11.41} \\
 & Neutral ID & \phasecell{phasePositive}{16}{black}{+6.53}{2.79} & \phasecell{phasePositive}{18}{black}{+8.10}{3.95} & \phasecell{phasePositive}{15}{black}{+6.12}{4.93} & \phasecell{phaseNegative}{17}{black}{-7.33}{3.41} & \phasecell{phasePositive}{9}{black}{+0.50}{2.02} & \phasecell{phaseNegative}{17}{black}{-7.32}{11.69} & \phasecell{phaseNegative}{17}{black}{-7.26}{6.94} & \phasecell{phaseNegative}{21}{black}{-10.87}{14.00} & \phasecell{phaseNegative}{24}{black}{-12.95}{9.25} \\
\midrule
Model & Qwen3-8B Base & \phasecell{phasePositive}{17}{black}{+7.17}{2.06} & \phasecell{phasePositive}{19}{black}{+9.09}{4.65} & \phasecell{phasePositive}{16}{black}{+6.44}{5.44} & \phasecell{phaseNegative}{18}{black}{-8.33}{3.33} & \phasecell{phaseNegative}{9}{black}{-0.73}{3.90} & \phasecell{phaseNegative}{12}{black}{-3.42}{7.82} & \phasecell{phasePositive}{23}{black}{+12.47}{9.96} & \phasecell{phasePositive}{32}{black}{+20.13}{10.57} & \phasecell{phasePositive}{40}{black}{+26.19}{9.52} \\
& Qwen3-8B Thinking & \phasecell{phasePositive}{12}{black}{+3.05}{1.87} & \phasecell{phasePositive}{14}{black}{+4.72}{2.96} & \phasecell{phasePositive}{11}{black}{+2.73}{5.33} & \phasecell{phaseNegative}{10}{black}{-1.66}{3.71} & \phasecell{phasePositive}{19}{black}{+9.38}{4.31} & \phasecell{phaseNegative}{19}{black}{-8.77}{13.15} & \phasecell{phaseNegative}{11}{black}{-2.11}{12.78} & \phasecell{phaseNegative}{33}{black}{-21.01}{12.32} & \phasecell{phaseNegative}{30}{black}{-18.32}{13.72} \\
& Qwen3-4B & \phasecell{phasePositive}{13}{black}{+4.17}{5.75} & \phasecell{phasePositive}{16}{black}{+6.30}{7.14} & \phasecell{phasePositive}{12}{black}{+3.45}{3.06} & \phasecell{phaseNegative}{14}{black}{-4.87}{6.37} & \phasecell{phaseNegative}{13}{black}{-4.07}{4.44} & \phasecell{phaseNegative}{24}{black}{-13.25}{12.15} & \phasecell{phaseNegative}{20}{black}{-10.33}{8.35} & \phasecell{phasePositive}{17}{black}{+7.48}{9.91} & \phasecell{phasePositive}{14}{black}{+5.34}{15.36} \\
& Gemma3-12B & \phasecell{phasePositive}{22}{black}{+11.46}{3.24} & \phasecell{phasePositive}{25}{black}{+13.79}{5.22} & \phasecell{phasePositive}{19}{black}{+8.73}{3.90} & \phasecell{phaseNegative}{24}{black}{-13.16}{5.08} & \phasecell{phaseNegative}{12}{black}{-2.98}{7.91} & \phasecell{phaseNegative}{25}{black}{-13.98}{14.62} & \phasecell{phaseNegative}{18}{black}{-7.93}{9.72} & \phasecell{phaseNegative}{26}{black}{-15.13}{20.28} & \phasecell{phaseNegative}{26}{black}{-14.96}{17.68} \\
& Ministral-3-8B & \phasecell{phasePositive}{9}{black}{+0.93}{2.12} & \phasecell{phasePositive}{9}{black}{+0.49}{1.37} & \phasecell{phaseNegative}{8}{black}{-0.22}{0.49} & \phasecell{phaseNegative}{9}{black}{-0.90}{2.15} & \phasecell{phasePositive}{9}{black}{+0.84}{1.50} & \phasecell{phasePositive}{9}{black}{+0.48}{0.37} & \phasecell{phasePositive}{9}{black}{+0.94}{1.91} & \phasecell{phasePositive}{12}{black}{+3.71}{7.47} & \phasecell{phaseNegative}{8}{black}{-0.16}{0.86} \\
& Gemma4-12B & \phasecell{phasePositive}{24}{black}{+13.40}{2.46} & \phasecell{phasePositive}{27}{black}{+15.49}{5.24} & \phasecell{phasePositive}{22}{black}{+11.92}{6.04} & \phasecell{phaseNegative}{26}{black}{-14.80}{4.27} & \phasecell{phaseNegative}{16}{black}{-7.00}{6.85} & \phasecell{phaseNegative}{33}{black}{-20.75}{16.52} & \phasecell{phaseNegative}{29}{black}{-17.60}{15.62} & \phasecell{phaseNegative}{30}{black}{-17.81}{21.88} & \phasecell{phaseNegative}{52}{black}{-36.50}{13.36} \\
& Granite-4-8B & \phasecell{phasePositive}{19}{black}{+9.17}{4.72} & \phasecell{phasePositive}{21}{black}{+10.97}{5.97} & \phasecell{phasePositive}{17}{black}{+7.83}{3.78} & \phasecell{phaseNegative}{21}{black}{-10.90}{6.56} & \phasecell{phaseNegative}{12}{black}{-3.33}{4.30} & \phasecell{phaseNegative}{29}{black}{-17.08}{12.96} & \phasecell{phaseNegative}{20}{black}{-10.29}{8.36} & \phasecell{phaseNegative}{10}{black}{-1.72}{7.84} & \phasecell{phasePositive}{21}{black}{+11.01}{17.71} \\
\midrule
Scale cardinality & 3-way & \phasecell{phasePositive}{38}{black}{+25.00}{6.17} & \phasecell{phasePositive}{37}{black}{+23.84}{6.20} & \phasecell{phasePositive}{11}{black}{+2.53}{4.30} & \phasecell{phaseNegative}{44}{black}{-29.42}{7.08} & \phasecell{phaseNegative}{28}{black}{-16.33}{0.34} & \phasecell{phaseNegative}{47}{black}{-32.37}{9.46} & \phasecell{phaseNegative}{40}{black}{-26.41}{3.74} & \phasecell{phaseNegative}{29}{black}{-17.22}{0.45} & \phasecell{phaseNegative}{39}{black}{-25.31}{4.51} \\
 & 2-way & \phasecell{phasePositive}{61}{white}{+43.85}{5.57} & \phasecell{phasePositive}{65}{white}{+47.47}{5.39} & \phasecell{phasePositive}{11}{black}{+2.47}{8.85} & \phasecell{phaseNegative}{68}{white}{-49.34}{6.90} & \phasecell{phaseNegative}{29}{black}{-17.34}{0.32} & \phasecell{phaseNegative}{65}{white}{-47.41}{8.21} & \phasecell{phaseNegative}{51}{black}{-35.57}{3.94} & \phasecell{phaseNegative}{48}{black}{-33.07}{3.85} & \phasecell{phaseNegative}{67}{white}{-48.91}{4.42} \\
\midrule
Demo count & 0-shot & \phasecell{phasePositive}{17}{black}{+7.13}{3.49} & \phasecell{phasePositive}{18}{black}{+8.01}{6.04} & \phasecell{phasePositive}{16}{black}{+6.45}{4.82} & \phasecell{phaseNegative}{17}{black}{-7.37}{4.74} & \phasecell{phasePositive}{16}{black}{+6.87}{4.11} & \phasena & \phasena & \phasena & \phasena \\
 & 1-shot & \phasecell{phasePositive}{13}{black}{+4.40}{3.63} & \phasecell{phasePositive}{13}{black}{+4.00}{5.69} & \phasecell{phasePositive}{15}{black}{+5.79}{4.63} & \phasecell{phaseNegative}{14}{black}{-4.83}{3.60} & \phasecell{phasePositive}{12}{black}{+3.43}{6.64} & \phasecell{phaseNegative}{28}{black}{-16.78}{14.23} & \phasecell{phaseNegative}{17}{black}{-7.33}{3.45} & \phasecell{phaseNegative}{16}{black}{-6.25}{11.80} & \phasecell{phaseNegative}{32}{black}{-20.16}{13.06} \\
 & 3-shot & \phasecell{phasePositive}{9}{black}{+0.77}{2.36} & \phasecell{phasePositive}{9}{black}{+0.80}{3.41} & \phasecell{phasePositive}{10}{black}{+1.78}{1.34} & \phasecell{phaseNegative}{10}{black}{-1.53}{2.10} & \phasecell{phaseNegative}{8}{black}{-0.13}{2.77} & \phasecell{phaseNegative}{19}{black}{-8.72}{10.49} & \phasecell{phaseNegative}{15}{black}{-6.20}{5.36} & \phasecell{phaseNegative}{26}{black}{-14.62}{12.19} & \phasecell{phaseNegative}{23}{black}{-12.09}{12.51} \\
\midrule
Prompt clarity & Minimal & \phasecell{phaseNegative}{12}{black}{-3.68}{17.52} & \phasecell{phaseNegative}{18}{black}{-8.06}{17.81} & \phasecell{phaseNegative}{39}{black}{-25.57}{19.42} & \phasecell{phasePositive}{29}{black}{+17.50}{32.40} & \phasecell{phasePositive}{67}{white}{+48.84}{27.54} & \phasecell{phaseNegative}{22}{black}{-11.89}{25.13} & \phasecell{phasePositive}{18}{black}{+8.43}{22.60} & \phasecell{phaseNegative}{24}{black}{-13.50}{22.02} & \phasecell{phaseNegative}{23}{black}{-12.21}{14.15} \\
 & Ordinal-explicit & \phasecell{phasePositive}{11}{black}{+2.77}{3.40} & \phasecell{phasePositive}{10}{black}{+1.60}{4.55} & \phasecell{phaseNegative}{11}{black}{-2.33}{8.14} & \phasecell{phaseNegative}{13}{black}{-4.10}{4.30} & \phasecell{phasePositive}{14}{black}{+4.97}{1.83} & \phasecell{phasePositive}{12}{black}{+3.48}{12.35} & \phasecell{phasePositive}{18}{black}{+8.54}{11.62} & \phasecell{phasePositive}{12}{black}{+3.38}{18.35} & \phasecell{phaseNegative}{16}{black}{-6.89}{1.69} \\
\midrule
Separator & Space & \phasecell{phaseNegative}{15}{black}{-5.50}{6.86} & \phasecell{phaseNegative}{16}{black}{-6.37}{8.03} & \phasecell{phaseNegative}{14}{black}{-5.07}{8.33} & \phasecell{phasePositive}{18}{black}{+7.95}{11.61} & \phasecell{phasePositive}{8}{black}{+0.22}{1.63} & \phasecell{phasePositive}{15}{black}{+5.38}{10.47} & \phasecell{phasePositive}{22}{black}{+11.20}{12.80} & \phasecell{phasePositive}{10}{black}{+1.72}{10.33} & \phasecell{phasePositive}{9}{black}{+0.93}{22.27} \\
 & Tab & \phasecell{phaseNegative}{13}{black}{-4.37}{8.84} & \phasecell{phaseNegative}{15}{black}{-5.81}{12.29} & \phasecell{phaseNegative}{16}{black}{-6.34}{16.49} & \phasecell{phasePositive}{22}{black}{+11.43}{22.99} & \phasecell{phasePositive}{8}{black}{+0.33}{5.99} & \phasecell{phasePositive}{20}{black}{+9.56}{18.80} & \phasecell{phasePositive}{27}{black}{+15.57}{17.49} & \phasecell{phaseNegative}{10}{black}{-1.59}{13.85} & \phasecell{phasePositive}{10}{black}{+1.55}{23.66} \\
\midrule
Connector & Space & \phasecell{phaseNegative}{18}{black}{-7.86}{8.53} & \phasecell{phaseNegative}{18}{black}{-8.17}{8.10} & \phasecell{phaseNegative}{19}{black}{-9.28}{8.27} & \phasecell{phasePositive}{22}{black}{+11.50}{10.25} & \phasecell{phasePositive}{9}{black}{+0.61}{2.92} & \phasecell{phasePositive}{16}{black}{+6.62}{7.51} & \phasecell{phasePositive}{27}{black}{+15.56}{9.57} & \phasecell{phaseNegative}{16}{black}{-6.23}{11.80} & \phasecell{phaseNegative}{10}{black}{-2.06}{12.43} \\
 & Newline+tab & \phasecell{phaseNegative}{12}{black}{-3.23}{2.78} & \phasecell{phaseNegative}{13}{black}{-3.91}{2.80} & \phasecell{phaseNegative}{12}{black}{-3.43}{3.22} & \phasecell{phasePositive}{13}{black}{+3.76}{2.61} & \phasecell{phasePositive}{10}{black}{+1.68}{2.55} & \phasecell{phasePositive}{9}{black}{+0.89}{0.58} & \phasecell{phasePositive}{12}{black}{+3.05}{2.26} & \phasecell{phaseNegative}{10}{black}{-2.02}{6.60} & \phasecell{phaseNegative}{16}{black}{-6.77}{4.28} \\
\midrule
Mood & Interrogative & \phasecell{phaseNegative}{12}{black}{-3.57}{4.46} & \phasecell{phaseNegative}{14}{black}{-5.37}{3.76} & \phasecell{phaseNegative}{14}{black}{-4.67}{5.63} & \phasecell{phasePositive}{12}{black}{+3.17}{4.55} & \phasecell{phasePositive}{13}{black}{+3.97}{3.78} & \phasecell{phasePositive}{9}{black}{+0.65}{2.79} & \phasecell{phasePositive}{33}{black}{+20.50}{16.07} & \phasecell{phaseNegative}{15}{black}{-6.14}{8.57} & \phasecell{phasePositive}{10}{black}{+1.82}{11.22} \\
 & Indicative & \phasecell{phasePositive}{10}{black}{+1.97}{1.61} & \phasecell{phasePositive}{11}{black}{+2.42}{2.11} & \phasecell{phasePositive}{10}{black}{+1.33}{1.76} & \phasecell{phaseNegative}{12}{black}{-2.90}{2.59} & \phasecell{phaseNegative}{8}{black}{-0.17}{2.08} & \phasecell{phaseNegative}{12}{black}{-3.65}{4.06} & \phasecell{phaseNegative}{12}{black}{-3.39}{4.07} & \phasecell{phasePositive}{14}{black}{+4.59}{3.67} & \phasecell{phasePositive}{11}{black}{+2.30}{1.52} \\
\bottomrule
\end{tabular}
\vspace{-2pt}
\caption{Mean Phase~1 factor residuals (all values $\times 100$). Each cell reports the factor-level value minus the baseline value, averaged over applicable datasets and demonstration seeds; full mean $\pm$ SD results appear in Supplementary Tables~\ref{tab:phase1_residuals_full_a} and~\ref{tab:phase1_residuals_full_b}.}
\label{tab:phase1_residuals}
\vspace{-7pt}
\end{table*}

%% file: sec/6_phase2.tex
\subsection{Phase 2: Pipeline and Joint-Factor Experiments}
\label{sec:phase2}

Phase~1 showed that local prompt and model choices substantially affect positional stability. Phase~2 examines whether broader interventions provide more reliable improvements. Tables~\ref{tab:priority_model_heatmap} and~\ref{tab:phase2_pipeline_residuals} report two complementary experiments: the former compares transfer models against Qwen3-8B under fixed configurations, while the latter evaluates alternative inference, aggregation, and debiasing pipelines relative to the Phase~1 baseline. Across both experiments, improvements remain method-, model-, and probe-specific.

\input{sec/6_priority_table}

\subsubsection{Phase 2a: Pipeline-Level Alternatives}

Table~\ref{tab:phase2_pipeline_residuals} reports pipeline residuals relative to the Phase~1 baseline. Inference paradigms and aggregation methods are averaged over the three non-binary datasets, whereas debiasing methods use all five. Because pairwise and listwise inference alter prompt structure, only label-order sensitivity is directly comparable across paradigms. Aggregation rows additionally report the positional probes derived from shared pairwise outputs. Label and demonstration averaging merge predictions across positional variants, leaving no comparable flip rate. Implementation details appear in Supplementary Section~\ref{app:phase2_config}.

The four debiasing strategies lead to two different conclusions. Label averaging and demonstration averaging leave predictive performance essentially unchanged: accuracy decreases by 1.1\% and 0.4\%, respectively. Because these methods merge predictions across positional variants, their final outputs do not retain a comparable flip-rate measure. Among the methods for which label-order sensitivity remains measurable, neither provides an improvement. PriDe~\citep{zheng2024large} increases accuracy by 0.9\% but also increases the label-order flip rate by 0.7\%. Contextual Calibration~\citep{zhao2021calibrate} lowers accuracy by 2.9\% and increases the flip rate by 3.3\%. Thus, averaging preserves performance, while neither calibration method reduces label-order bias.

Changing the inference paradigm has much larger effects than applying a debiasing correction. The default pairwise pipeline worsens both goals: accuracy decreases by 8.9\%, while the label-order flip rate increases by 27.8\%. Changing the aggregation rule shifts this trade-off rather than removing it. Copeland preserves accuracy but still increases label-order flips by 24.5\%. Mean-score aggregation reduces the flip rate slightly, but lowers accuracy by 16.3\%. Bradley--Terry performs poorly on both measures, reducing accuracy by 20.4\% and increasing label-order flips by 43.0\%. Thus, none of the tested aggregation rules makes pairwise inference both accurate and stable.

The listwise results show that the exact formulation matters more than the broad paradigm name. Direct listwise inference causes the largest performance loss: accuracy decreases by 27.6\%, Spearman correlation falls by 72.6\%, and MAE increases by 52.1\%, while label-order sensitivity barely changes. Listwise-compare produces the opposite pattern. Accuracy remains within 0.7\% of the baseline, F1 and Spearman correlation improve slightly, and the label-order flip rate falls by 18.1\%. Listwise-compare is therefore a promising pipeline, but its success should not be interpreted as a general benefit of listwise prompting.

Taken together, Table~\ref{tab:phase2_pipeline_residuals} shows that changing the pipeline is not automatically beneficial. Averaging preserves performance but leaves no directly comparable residual flip rate; PriDe and Contextual Calibration do not improve label-order stability; and the default pairwise and direct listwise pipelines sacrifice performance. Listwise-compare is the only balanced improvement in this experiment.

\subsubsection{Phase 2b: Combining Phase~1 Factors}

Table~\ref{tab:priority_model_heatmap} asks whether a configuration selected on Qwen3 produces a similar metric profile on other models. We construct four configurations using Qwen3-8B Instruct: \emph{A} targets accuracy, \emph{P1} targets label-order stability, and \emph{P2a} and \emph{P2b} target the two demonstration-order contrasts. The complete configurations appear in Supplementary Section~\ref{app:priority_config}. Each cell subtracts the Qwen3 value under the same configuration from the transfer-model value; positive residuals therefore indicate larger values than Qwen3, which is beneficial for accuracy, F1, and $\rho$ but harmful for MAE and PFR.

The accuracy-oriented configuration transfers unevenly. Relative to Qwen3 under \emph{A}, accuracy is 7.1 points lower on Ministral, 2.3 points lower on Gemma3, and 9.6 points lower on Granite. Label-order PFR is 17.1 points higher on Ministral but 6.8 and 3.5 points lower on Gemma3 and Granite. Thus, even when the same configuration is applied, both performance and stability can shift in different directions across models.

The label-order configuration shows the same lack of uniformity. Its target PFR is 5.4 points higher than Qwen3 on Ministral and 2.9 points higher on Gemma3, but 2.8 points lower on Granite. Accuracy ranges from 3.5 points below Qwen3 on Ministral to 8.7 and 5.3 points above it on Gemma3 and Granite. Optimizing the target probe on Qwen3 therefore does not define a model-independent operating point.

The demonstration-order configurations remain metric-specific. Under \emph{P2a}, the target P2a residual ranges from $+10.3$ points on Ministral to $-4.2$ on Granite, while label-order PFR remains close to the already high Qwen3 value for all three transfer models. Under \emph{P2b}, target residuals are smaller ($+3.4$, $+1.6$, and $-2.5$ points), but both placement PFRs are higher than Qwen3 on every transfer model. A configuration can therefore resemble Qwen3 on its target probe while diverging sharply on another positional test.

Taken together, Table~\ref{tab:priority_model_heatmap} shows that joint configurations are model- and objective-specific. Cross-model validation should compare the complete performance and stability profile rather than assume that a Qwen-selected configuration transfers as a fixed recipe.

\FloatBarrier

%% file: sec/6_priority_table.tex
\definecolor{priorityPositive}{HTML}{C65F5F}
\definecolor{priorityNegative}{HTML}{4C78A8}
\newcommand{\prioritycell}[5]{\colorbox{#1!#2}{\textcolor{#3}{\makebox[3.70em][c]{$#4$}}}}
\newcommand{\priorityna}{\colorbox{gray!20}{\makebox[3.70em][c]{--}}}

\begin{table*}[!t]
\centering
\small
\setlength{\tabcolsep}{0.5pt}
\renewcommand{\arraystretch}{0.92}
\setlength{\aboverulesep}{0pt}
\setlength{\belowrulesep}{0pt}
\begin{tabular}{l@{\hspace{4pt}}lccccccccc}
\toprule
& & \multicolumn{4}{c}{Performance residuals} & \multicolumn{5}{c}{Position-sensitivity residuals} \\
Model & Config. & Acc.$\uparrow$ & F1$\uparrow$ & $\rho\uparrow$ & MAE$\downarrow$ & P1$\downarrow$ & P2a$\downarrow$ & P2b$\downarrow$ & P3a$\downarrow$ & P3b$\downarrow$ \\
\midrule
Ministral-3-8B & baseline & \prioritycell{priorityPositive}{9}{black}{+0.90}{0.00} & \prioritycell{priorityPositive}{9}{black}{+0.50}{0.00} & \prioritycell{priorityNegative}{8}{black}{-0.30}{0.00} & \prioritycell{priorityNegative}{9}{black}{-0.90}{0.00} & \prioritycell{priorityPositive}{9}{black}{+0.90}{0.00} & \prioritycell{priorityPositive}{8}{black}{+0.40}{0.00} & \prioritycell{priorityPositive}{9}{black}{+0.90}{0.00} & \prioritycell{priorityPositive}{12}{black}{+3.70}{0.00} & \prioritycell{priorityNegative}{8}{black}{-0.20}{0.00} \\
 & A & \prioritycell{priorityNegative}{17}{black}{-7.10}{0.00} & \prioritycell{priorityNegative}{19}{black}{-9.30}{0.00} & \prioritycell{priorityNegative}{15}{black}{-6.20}{0.00} & \prioritycell{priorityPositive}{17}{black}{+7.10}{0.00} & \prioritycell{priorityPositive}{29}{black}{+17.10}{0.00} & \priorityna & \priorityna & \priorityna & \priorityna \\
 & P1 & \prioritycell{priorityNegative}{12}{black}{-3.50}{0.00} & \prioritycell{priorityNegative}{12}{black}{-3.00}{0.00} & \prioritycell{priorityNegative}{13}{black}{-4.40}{0.00} & \prioritycell{priorityPositive}{15}{black}{+5.60}{0.00} & \prioritycell{priorityPositive}{14}{black}{+5.40}{0.00} & \prioritycell{priorityPositive}{9}{black}{+0.50}{0.00} & \prioritycell{priorityPositive}{18}{black}{+8.00}{0.00} & \prioritycell{priorityNegative}{9}{black}{-1.20}{0.00} & \prioritycell{priorityNegative}{9}{black}{-1.20}{0.00} \\
 & P2a & \prioritycell{priorityNegative}{21}{black}{-10.70}{0.00} & \prioritycell{priorityNegative}{22}{black}{-11.70}{0.00} & \prioritycell{priorityNegative}{21}{black}{-10.60}{0.00} & \prioritycell{priorityPositive}{23}{black}{+12.30}{0.00} & \prioritycell{priorityPositive}{9}{black}{+0.70}{0.00} & \prioritycell{priorityPositive}{20}{black}{+10.30}{0.00} & \prioritycell{priorityPositive}{15}{black}{+6.20}{0.00} & \prioritycell{priorityNegative}{13}{black}{-4.10}{0.00} & \prioritycell{priorityNegative}{9}{black}{-1.20}{0.00} \\
 & P2b & \prioritycell{priorityNegative}{12}{black}{-3.30}{0.00} & \prioritycell{priorityNegative}{11}{black}{-2.60}{0.00} & \prioritycell{priorityNegative}{12}{black}{-3.30}{0.00} & \prioritycell{priorityPositive}{13}{black}{+3.90}{0.00} & \prioritycell{priorityNegative}{19}{black}{-9.50}{0.00} & \prioritycell{priorityPositive}{12}{black}{+3.70}{0.00} & \prioritycell{priorityPositive}{12}{black}{+3.40}{0.00} & \prioritycell{priorityPositive}{20}{black}{+10.20}{0.00} & \prioritycell{priorityPositive}{12}{black}{+3.10}{0.00} \\
\midrule
Gemma3-12B & baseline & \prioritycell{priorityPositive}{22}{black}{+11.40}{0.00} & \prioritycell{priorityPositive}{25}{black}{+13.80}{0.00} & \prioritycell{priorityPositive}{18}{black}{+8.70}{0.00} & \prioritycell{priorityNegative}{24}{black}{-13.20}{0.00} & \prioritycell{priorityNegative}{11}{black}{-2.90}{0.00} & \prioritycell{priorityNegative}{25}{black}{-14.00}{0.00} & \prioritycell{priorityNegative}{18}{black}{-8.00}{0.00} & \prioritycell{priorityNegative}{26}{black}{-15.10}{0.00} & \prioritycell{priorityNegative}{26}{black}{-15.00}{0.00} \\
 & A & \prioritycell{priorityNegative}{11}{black}{-2.30}{0.00} & \prioritycell{priorityNegative}{11}{black}{-2.50}{0.00} & \prioritycell{priorityNegative}{8}{black}{-0.30}{0.00} & \prioritycell{priorityPositive}{9}{black}{+1.10}{0.00} & \prioritycell{priorityNegative}{16}{black}{-6.80}{0.00} & \priorityna & \priorityna & \priorityna & \priorityna \\
 & P1 & \prioritycell{priorityPositive}{18}{black}{+8.70}{0.00} & \prioritycell{priorityPositive}{20}{black}{+10.30}{0.00} & \prioritycell{priorityPositive}{16}{black}{+6.90}{0.00} & \prioritycell{priorityNegative}{19}{black}{-9.10}{0.00} & \prioritycell{priorityPositive}{11}{black}{+2.90}{0.00} & \prioritycell{priorityNegative}{13}{black}{-4.50}{0.00} & \prioritycell{priorityNegative}{10}{black}{-1.40}{0.00} & \prioritycell{priorityPositive}{29}{black}{+17.80}{0.00} & \prioritycell{priorityPositive}{22}{black}{+11.60}{0.00} \\
 & P2a & \prioritycell{priorityPositive}{10}{black}{+1.50}{0.00} & \prioritycell{priorityPositive}{9}{black}{+1.00}{0.00} & \prioritycell{priorityPositive}{11}{black}{+2.40}{0.00} & \prioritycell{priorityNegative}{11}{black}{-2.40}{0.00} & \prioritycell{priorityPositive}{8}{black}{+0.40}{0.00} & \prioritycell{priorityPositive}{10}{black}{+1.50}{0.00} & \prioritycell{priorityNegative}{11}{black}{-2.20}{0.00} & \prioritycell{priorityNegative}{18}{black}{-8.00}{0.00} & \prioritycell{priorityNegative}{22}{black}{-11.30}{0.00} \\
 & P2b & \prioritycell{priorityPositive}{13}{black}{+4.30}{0.00} & \prioritycell{priorityPositive}{14}{black}{+4.80}{0.00} & \prioritycell{priorityPositive}{11}{black}{+2.60}{0.00} & \prioritycell{priorityNegative}{14}{black}{-4.90}{0.00} & \prioritycell{priorityNegative}{13}{black}{-4.20}{0.00} & \prioritycell{priorityPositive}{11}{black}{+2.20}{0.00} & \prioritycell{priorityPositive}{10}{black}{+1.60}{0.00} & \prioritycell{priorityPositive}{33}{black}{+21.10}{0.00} & \prioritycell{priorityPositive}{20}{black}{+9.60}{0.00} \\
\midrule
Granite-4-8B & baseline & \prioritycell{priorityPositive}{19}{black}{+9.10}{0.00} & \prioritycell{priorityPositive}{21}{black}{+11.00}{0.00} & \prioritycell{priorityPositive}{17}{black}{+7.80}{0.00} & \prioritycell{priorityNegative}{21}{black}{-10.90}{0.00} & \prioritycell{priorityNegative}{12}{black}{-3.30}{0.00} & \prioritycell{priorityNegative}{29}{black}{-17.10}{0.00} & \prioritycell{priorityNegative}{20}{black}{-10.30}{0.00} & \prioritycell{priorityNegative}{10}{black}{-1.70}{0.00} & \prioritycell{priorityPositive}{21}{black}{+11.00}{0.00} \\
 & A & \prioritycell{priorityNegative}{20}{black}{-9.60}{0.00} & \prioritycell{priorityNegative}{19}{black}{-9.50}{0.00} & \prioritycell{priorityNegative}{26}{black}{-15.40}{0.00} & \prioritycell{priorityPositive}{19}{black}{+8.90}{0.00} & \prioritycell{priorityNegative}{12}{black}{-3.50}{0.00} & \priorityna & \priorityna & \priorityna & \priorityna \\
 & P1 & \prioritycell{priorityPositive}{14}{black}{+5.30}{0.00} & \prioritycell{priorityPositive}{15}{black}{+5.90}{0.00} & \prioritycell{priorityPositive}{14}{black}{+4.80}{0.00} & \prioritycell{priorityNegative}{15}{black}{-5.60}{0.00} & \prioritycell{priorityNegative}{11}{black}{-2.80}{0.00} & \prioritycell{priorityNegative}{21}{black}{-10.90}{0.00} & \prioritycell{priorityNegative}{14}{black}{-5.30}{0.00} & \prioritycell{priorityPositive}{23}{black}{+12.50}{0.00} & \prioritycell{priorityPositive}{27}{black}{+16.20}{0.00} \\
 & P2a & \prioritycell{priorityNegative}{11}{black}{-2.90}{0.00} & \prioritycell{priorityNegative}{15}{black}{-5.60}{0.00} & \prioritycell{priorityNegative}{13}{black}{-4.50}{0.00} & \prioritycell{priorityPositive}{12}{black}{+3.30}{0.00} & \prioritycell{priorityPositive}{10}{black}{+2.00}{0.00} & \prioritycell{priorityNegative}{13}{black}{-4.20}{0.00} & \prioritycell{priorityNegative}{14}{black}{-4.80}{0.00} & \prioritycell{priorityNegative}{15}{black}{-6.10}{0.00} & \prioritycell{priorityPositive}{17}{black}{+7.40}{0.00} \\
 & P2b & \prioritycell{priorityPositive}{10}{black}{+1.80}{0.00} & \prioritycell{priorityPositive}{11}{black}{+2.20}{0.00} & \prioritycell{priorityPositive}{10}{black}{+1.60}{0.00} & \prioritycell{priorityNegative}{10}{black}{-2.00}{0.00} & \prioritycell{priorityPositive}{9}{black}{+0.80}{0.00} & \prioritycell{priorityPositive}{8}{black}{+0.30}{0.00} & \prioritycell{priorityNegative}{11}{black}{-2.50}{0.00} & \prioritycell{priorityPositive}{21}{black}{+10.50}{0.00} & \prioritycell{priorityPositive}{32}{black}{+20.30}{0.00} \\
\bottomrule
\end{tabular}
\vspace{-2pt}
\caption{Cross-model Phase~2 residuals (all values $\times 100$). Each cell reports the transfer-model value minus the Qwen3-8B value under the same configuration; underlying absolute mean $\pm$ SD results appear in Supplementary Table~\ref{tab:phase2_joint_full}.}
\label{tab:priority_model_heatmap}
\vspace{-5pt}
\end{table*}

\begin{table*}[!t]
\centering
\small
\setlength{\tabcolsep}{0.5pt}
\renewcommand{\arraystretch}{0.92}
\setlength{\aboverulesep}{0pt}
\setlength{\belowrulesep}{0pt}
\begin{tabular}{l@{\hspace{4pt}}lccccccccc}
\toprule
& & \multicolumn{4}{c}{Performance residuals} & \multicolumn{5}{c}{Position-sensitivity residuals} \\
Family & Method & Acc.$\uparrow$ & F1$\uparrow$ & $\rho\uparrow$ & MAE$\downarrow$ & P1$\downarrow$ & P2a$\downarrow$ & P2b$\downarrow$ & P3a$\downarrow$ & P3b$\downarrow$ \\
\midrule
Inference paradigm & Pairwise & \prioritycell{priorityNegative}{17}{black}{-8.94}{2.86} & \prioritycell{priorityNegative}{18}{black}{-9.70}{3.73} & \prioritycell{priorityNegative}{14}{black}{-6.23}{3.57} & \prioritycell{priorityPositive}{25}{black}{+17.05}{2.42} & \prioritycell{priorityPositive}{36}{black}{+27.83}{12.75} & \priorityna & \priorityna & \priorityna & \priorityna \\
& Listwise & \prioritycell{priorityNegative}{35}{black}{-27.61}{3.89} & \prioritycell{priorityNegative}{43}{black}{-35.10}{4.00} & \prioritycell{priorityNegative}{80}{white}{-72.63}{14.13} & \prioritycell{priorityPositive}{60}{white}{+52.06}{11.11} & \prioritycell{priorityPositive}{10}{black}{+2.11}{31.87} & \priorityna & \priorityna & \priorityna & \priorityna \\
& Listwise-compare & \prioritycell{priorityNegative}{9}{black}{-0.67}{7.05} & \prioritycell{priorityPositive}{9}{black}{+0.55}{5.84} & \prioritycell{priorityPositive}{14}{black}{+5.81}{3.79} & \prioritycell{priorityPositive}{10}{black}{+2.44}{9.82} & \prioritycell{priorityNegative}{26}{black}{-18.11}{6.64} & \priorityna & \priorityna & \priorityna & \priorityna \\
\midrule
Aggregation method & Majority vote & \prioritycell{priorityNegative}{15}{black}{-7.22}{0.00} & \prioritycell{priorityNegative}{13}{black}{-5.46}{0.00} & \prioritycell{priorityNegative}{25}{black}{-17.61}{0.00} & \prioritycell{priorityPositive}{26}{black}{+18.39}{0.00} & \prioritycell{priorityPositive}{44}{black}{+35.89}{0.00} & \prioritycell{priorityNegative}{38}{black}{-30.58}{0.00} & \prioritycell{priorityNegative}{30}{black}{-22.44}{0.00} & \prioritycell{priorityNegative}{75}{white}{-67.43}{0.00} & \prioritycell{priorityNegative}{64}{white}{-56.10}{0.00} \\
& Mean score & \prioritycell{priorityNegative}{24}{black}{-16.28}{0.00} & \prioritycell{priorityNegative}{31}{black}{-23.18}{0.00} & \prioritycell{priorityNegative}{27}{black}{-18.84}{0.00} & \prioritycell{priorityPositive}{32}{black}{+23.83}{0.00} & \prioritycell{priorityNegative}{9}{black}{-1.39}{0.00} & \prioritycell{priorityNegative}{52}{black}{-44.69}{0.00} & \prioritycell{priorityNegative}{44}{black}{-36.27}{0.00} & \prioritycell{priorityNegative}{75}{white}{-67.43}{0.00} & \prioritycell{priorityNegative}{64}{white}{-56.10}{0.00} \\
& Weighted sum & \prioritycell{priorityNegative}{17}{black}{-8.94}{0.00} & \prioritycell{priorityNegative}{18}{black}{-9.71}{0.00} & \prioritycell{priorityNegative}{14}{black}{-6.27}{0.00} & \prioritycell{priorityPositive}{25}{black}{+17.17}{0.00} & \prioritycell{priorityPositive}{35}{black}{+27.22}{0.00} & \prioritycell{priorityNegative}{49}{black}{-40.97}{0.00} & \prioritycell{priorityNegative}{40}{black}{-32.05}{0.00} & \prioritycell{priorityNegative}{75}{white}{-67.43}{0.00} & \prioritycell{priorityNegative}{64}{white}{-56.10}{0.00} \\
& Copeland & \prioritycell{priorityNegative}{9}{black}{-0.61}{0.00} & \prioritycell{priorityPositive}{8}{black}{+0.47}{0.00} & \prioritycell{priorityNegative}{10}{black}{-2.51}{0.00} & \prioritycell{priorityPositive}{10}{black}{+1.89}{0.00} & \prioritycell{priorityPositive}{32}{black}{+24.50}{0.00} & \prioritycell{priorityNegative}{49}{black}{-41.25}{0.00} & \prioritycell{priorityNegative}{40}{black}{-32.55}{0.00} & \prioritycell{priorityNegative}{75}{white}{-67.43}{0.00} & \prioritycell{priorityNegative}{64}{white}{-56.10}{0.00} \\
& Bradley--Terry & \prioritycell{priorityNegative}{28}{black}{-20.45}{0.00} & \prioritycell{priorityNegative}{31}{black}{-23.46}{0.00} & \prioritycell{priorityNegative}{50}{black}{-42.12}{0.00} & \prioritycell{priorityPositive}{73}{white}{+66.00}{0.00} & \prioritycell{priorityPositive}{51}{black}{+43.00}{0.00} & \prioritycell{priorityNegative}{22}{black}{-13.97}{0.00} & \prioritycell{priorityNegative}{21}{black}{-13.56}{0.00} & \prioritycell{priorityNegative}{75}{white}{-67.43}{0.00} & \prioritycell{priorityNegative}{64}{white}{-56.10}{0.00} \\
\midrule
Debiasing strategy & Label averaging & \prioritycell{priorityNegative}{9}{black}{-1.07}{1.92} & \prioritycell{priorityNegative}{10}{black}{-1.63}{1.96} & \prioritycell{priorityNegative}{9}{black}{-1.04}{2.33} & \prioritycell{priorityPositive}{9}{black}{+0.53}{2.31} & \priorityna & \priorityna & \priorityna & \priorityna & \priorityna \\
& Demo averaging & \prioritycell{priorityNegative}{8}{black}{-0.42}{5.23} & \prioritycell{priorityPositive}{8}{black}{+0.20}{6.32} & \prioritycell{priorityNegative}{9}{black}{-0.61}{2.64} & \prioritycell{priorityPositive}{9}{black}{+0.83}{5.38} & \priorityna & \priorityna & \priorityna & \priorityna & \priorityna \\
& PriDe & \prioritycell{priorityPositive}{9}{black}{+0.90}{1.27} & \prioritycell{priorityPositive}{10}{black}{+2.11}{2.10} & \prioritycell{priorityPositive}{9}{black}{+1.38}{2.29} & \prioritycell{priorityNegative}{9}{black}{-0.60}{1.26} & \prioritycell{priorityPositive}{9}{black}{+0.73}{1.48} & \priorityna & \priorityna & \priorityna & \priorityna \\
& Context. Calib. & \prioritycell{priorityNegative}{11}{black}{-2.87}{4.69} & \prioritycell{priorityNegative}{11}{black}{-3.27}{5.74} & \prioritycell{priorityNegative}{9}{black}{-1.37}{3.55} & \prioritycell{priorityPositive}{13}{black}{+5.13}{7.09} & \prioritycell{priorityPositive}{11}{black}{+3.30}{4.59} & \priorityna & \priorityna & \priorityna & \priorityna \\
\bottomrule
\end{tabular}
\vspace{-2pt}
\caption{Phase~2 pipeline residuals relative to the Qwen3-8B Phase~1 baseline (all values $\times 100$). The table compares inference paradigms, pairwise aggregation methods, and debiasing strategies; full mean $\pm$ SD results appear in Supplementary Table~\ref{tab:phase2_pipeline_full}.}
\label{tab:phase2_pipeline_residuals}
\vspace{-7pt}
\end{table*}

%% file: sec/7_conclusion.tex
\section{Conclusion}
We measured position bias in LLM-based ordinal classification across models, prompts, and pipelines. Accuracy and consistency respond to different choices; no tested correction eliminated bias, and favorable configurations transferred unevenly. By systematically separating three positional effects from predictive performance, this study shows that robustness belongs to the full inference configuration rather than to the model alone. These findings make positional consistency essential for evaluating LLM-based ordinal classifiers. Future work will examine factor interactions, asymmetric bias, and training-time interventions. \cleardoublepage

%% file: appendix/appendix.tex
\section{Phase 0 Model and API Configuration}
\label{app:phase0_api}

All Phase~0 requests were sent through the OpenRouter gateway at \url{https://openrouter.ai/api/v1} using the OpenAI-compatible \texttt{chat.completions.create} interface, corresponding to \texttt{/api/v1/chat/completions}. We did not set provider preferences or use routing suffixes such as \texttt{:nitro} or \texttt{:floor}; provider selection and any fallback were therefore handled by OpenRouter's default routing for each model ID. Table~\ref{tab:phase0_api_models} lists the exact model identifiers, access dates, response limits, and reasoning overrides.

\begin{table*}[t]
\centering
\footnotesize
\setlength{\tabcolsep}{2pt}
\begin{tabular}{@{}>{\raggedright\arraybackslash}p{0.17\textwidth}>{\raggedright\arraybackslash}p{0.09\textwidth}>{\raggedright\arraybackslash}p{0.30\textwidth}c c>{\raggedright\arraybackslash}p{0.12\textwidth}@{}}
\toprule
Model & Provider & OpenRouter model ID & Date (UTC) & Max tokens & Reasoning effort \\
\midrule
Claude Haiku 4.5 & Anthropic & \path{anthropic/claude-haiku-4.5} & 2026-06-16 & 512 & -- \\
Claude Sonnet 4.5 & Anthropic & \path{anthropic/claude-sonnet-4.5} & 2026-06-16 & 512 & -- \\
GPT-5.4 & OpenAI & \path{openai/gpt-5.4} & 2026-06-16 & 512 & -- \\
Gemini 3.1 Flash-Lite & Google & \path{google/gemini-3.1-flash-lite} & 2026-06-16 & 512 & -- \\
Gemini 3.1 Pro Preview & Google & \path{google/gemini-3.1-pro-preview} & 2026-06-17 & 2048 & Minimal \\
DeepSeek V4 Flash & DeepSeek & \path{deepseek/deepseek-v4-flash} & 2026-06-17 & 2048 & Minimal \\
DeepSeek V4 Pro & DeepSeek & \path{deepseek/deepseek-v4-pro} & 2026-06-17 & 2048 & Minimal \\
Grok 4.3 & xAI & \path{x-ai/grok-4.3} & 2026-06-16 & 512 & -- \\
Llama 4 Maverick & Meta & \path{meta-llama/llama-4-maverick} & 2026-06-16 & 512 & -- \\
Qwen3.7-Plus & Alibaba & \path{qwen/qwen3.7-plus} & 2026-06-16 & 512 & -- \\
\bottomrule
\end{tabular}
\caption{Phase~0 OpenRouter models and per-model request overrides.}
\label{tab:phase0_api_models}
\parbox{0.94\textwidth}{\small\textit{Note.} Minimal denotes \texttt{reasoning: \{"effort": "minimal"\}}; a dash indicates that no reasoning override was sent.}
\end{table*}

Table~\ref{tab:phase0_decoding} reports the decoding parameters. Phase~0a fixes the sampling seed and restricts decoding to the highest-probability token, whereas Phase~0b leaves the seed and \texttt{top\_k} unset. Parameters not shown were left at the OpenRouter or provider defaults.

\begin{table}[t]
\centering
\small
\begin{tabular}{lcc}
\toprule
Parameter & Phase~0a & Phase~0b \\
\midrule
\texttt{temperature} & 0 & 0.7 \\
\texttt{top\_k} & 1 & Not set \\
\texttt{top\_p} & 1.0 & 1.0 \\
\texttt{min\_p} & 0.0 & 0.0 \\
\texttt{seed} & 42 & Not set \\
\texttt{repeat\_penalty} & 1.0 & 1.0 \\
\bottomrule
\end{tabular}
\caption{Phase~0 decoding settings.}
\label{tab:phase0_decoding}
\end{table}

Requests were separated by 0.5 seconds. The supplied run configuration records no custom client-side retry policy. Prompts instructed the model to return only one numeric label. Outputs were mapped to a prediction only when the parser recovered a valid label for the task; reversed-label outputs were then mapped back to the natural semantic class space, and outputs without a valid label were counted as parse failures. Phase~0b uses independently sampled stochastic runs on the same fixed 200 test instances. For each instance, we form a noise-flip indicator from two runs of the same prompt and a total-flip indicator from the natural- and reversed-label runs, then compare these paired indicators using a one-sided McNemar test.

\section{Hidden Reasoning Tokens}
\label{app:hidden_tokens}

Table~\ref{tab:hidden_tokens} reports hidden reasoning tokens detected in Phase~0b. All models produce a single visible character, but total output tokens vary widely, indicating substantial hidden generation for some models despite the label-only output constraint and the minimal-reasoning override used where specified above. Gemini 3.1 Pro is an exception: despite generating 126 hidden tokens, its noise PFR remains low (0.045), suggesting that its internal reasoning may operate more deterministically. Requesting minimal reasoning or a one-character visible answer therefore does not guarantee the absence of hidden tokens, and researchers should monitor the gap between total output tokens and visible response length.

\begin{table*}[t]
\centering
\small
\begin{tabular}{lccc}
\toprule
Model & Output tokens\textsuperscript{a} & Hidden tokens & Noise PFR \\
\midrule
Gemini 3.1 Flash-Lite & 1 & 0 & 0.060 \\
Llama 4 Maverick & 2 & 1 & 0.060 \\
Claude Haiku 4.5 & 5 & 4 & 0.050 \\
Claude Sonnet 4.5 & 5 & 4 & 0.000 \\
GPT-5.4 & 5 & 4 & 0.050 \\
Gemini 3.1 Pro & 127 & 126 & 0.045 \\
DeepSeek V4 Flash & 157 & 156 & 0.130 \\
DeepSeek V4 Pro & 269 & 268 & 0.160 \\
Grok 4.3 & 333 & 332 & 0.075 \\
Qwen3.7-Plus & 513 & 512 & 0.135 \\
\bottomrule
\end{tabular}
\caption{Hidden reasoning tokens in Phase~0b.}
\label{tab:hidden_tokens}
\parbox{0.82\textwidth}{\small\textit{Note.} \textsuperscript{a}Output tokens are the average total reported by OpenRouter per response. All visible responses contain a single character; hidden tokens are the remaining output tokens.}
\end{table*}

\section{Dataset Access and Licensing}
\label{app:dataset_access}

Table~\ref{tab:dataset_access} records the official access route and stated usage terms for each dataset, as checked on July~28, 2026. We do not infer a license when the official release page does not state one. The Twitter-derived datasets may contain sensitive or offensive user-generated content; our paper reports only aggregate results and does not reproduce usernames or tweet text.

\begin{table*}[t]
\centering
\small
\setlength{\tabcolsep}{3pt}
\begin{tabular}{@{}>{\raggedright\arraybackslash}p{0.10\textwidth}>{\raggedright\arraybackslash}p{0.43\textwidth}>{\raggedright\arraybackslash}p{0.41\textwidth}@{}}
\toprule
Dataset & Official access & License or usage terms \\
\midrule
SST-5 & Stanford Sentiment Treebank download page: \url{https://nlp.stanford.edu/sentiment/code.html} & Publicly downloadable; the official dataset page does not state a separate dataset license. Users should retain the original citation and consult the current Stanford distribution terms. \\
Yelp-5 & Benchmark files from Zhang et al.~\citeyearpar{zhang2015character}, derived from the Yelp Dataset Challenge 2015; current portal: \url{https://business.yelp.com/data/resources/open-dataset/} & Yelp's Dataset Terms of Use govern access and restrict the Open Dataset to academic purposes. \\
Twitter & SemEval-2017 Task~4 data portal: \url{https://alt.qcri.org/semeval2017/task4/} & Use is subject to Twitter's platform terms; the task organizers require users to remove tweets that are no longer public. \\
Hate & HatEval official page and linked repository: \url{https://hatespeech.di.unito.it/hateval.html} & Creative Commons Attribution--NonCommercial 4.0 (CC BY-NC 4.0). \\
Offensive & OLID/OffensEval release page: \url{https://sites.google.com/site/offensevalsharedtask/olid} & Publicly released for the shared task; the cited release materials do not state a separate dataset license. Use of tweet content remains subject to the applicable platform terms. \\
\bottomrule
\end{tabular}
\caption{Dataset access and usage terms.}
\label{tab:dataset_access}
\end{table*}

\section{Baseline Configuration}
\label{app:baseline}

Table~\ref{tab:base_setting} lists the baseline configuration from which all factor ablations depart. We briefly justify the key choices. We select Qwen3-8B Instruct as the base model because the Qwen3 family offers base, instruct, and thinking variants on the same architecture, enabling controlled comparisons across training paradigms in the model-choice analysis. The 8B scale is large enough to perform competitively on ordinal classification while remaining affordable to run across all conditions. We use all five datasets rather than a single one so that factor levels can be evaluated across different domains and scale cardinalities where applicable, avoiding conclusions that are artifacts of one dataset. The remaining baseline dimensions follow standard practice in the few-shot ICL literature: numeric label format, 5-shot per class, demonstrations placed before the query, explicit imperative instructions, colon-plus-space separators, newline-connected fields, and greedy decoding for deterministic outputs. Following LAMPO, we evaluate three demonstration seeds~\citep{qin2024lampo}; our experiments use seed IDs 0, 1, and 42 with fixed 200-instance stratified test subsets. For decoding, we use \texttt{do\_sample = False} to ensure reproducible outputs and set a maximum response length to simplify parsing, while prompting the model to return a single-token answer.

The local-model experiments in Phases~1 and 2 were run on a host equipped with a 16-core CPU, 64~GB of system memory, and an NVIDIA L40S GPU. Phase~0 used models hosted through OpenRouter; the providers' server-side CPU, GPU, and memory configurations were not observable.

\begin{table*}[t]
\centering
\small
\begin{tabular}{lp{0.68\textwidth}}
\toprule
Dimension & Base value \\
\midrule
Model & Qwen3-8B Instruct (non-thinking) \\
Datasets & SST-5, Yelp-5, Twitter, Hate, Offensive \\
Label format & Numeric (1/2/3/4/5) \\
Inference paradigm & Pointwise ICL \\
Few-shot count & 5 per class \\
Label order & Natural (1 = most negative $\rightarrow$ 5 = most positive) \\
Debiasing & None \\
Demo order & Base \\
Demo placement & All before query \\
Prompt clarity & Explicit \\
Separator & Colon followed by a space \\
Connector & Newline \\
Instruction mood & Imperative \\
Seeds & (0, 1, 42) \\
Test instances & 200 per dataset (stratified) \\
Decoding & Greedy (\texttt{do\_sample = False}) \\
Response Length & \texttt{max\_new\_tokens = 32}\\

\bottomrule
\end{tabular}
\caption{Baseline configuration.}
\label{tab:base_setting}
\parbox{0.82\textwidth}{\small\textit{Note.} Phase~1 and Phase~2 experiments use these settings unless a factor or pipeline explicitly replaces them.}
\end{table*}

Here, we further specify how the baseline experiment instantiates the prompt; later ablation prompts build on it:

\begin{lstlisting}[frame=single,basicstyle=\ttfamily\footnotesize,breaklines=true,breakautoindent=false,breakindent=0pt]
Please perform Sentiment Classification task.
Given the sentence, assign a label from [1: very negative, 2: negative, 3: neutral, 4: positive, 5: very positive].
Return label only without any other text.

Sentence: ... a sweetly affecting story about four sisters who are coping , in one way or another , with life 's endgame .
Label: 4

Sentence: ` Like a child with an important message to tell ... ( Skins ' ) faults are easy to forgive because the intentions are lofty . '
Label: 3

[Remaining 23 demos with random order and each label has 5 demos total]

Sentence: Everywhere the camera looks there is something worth seeing .
Label:
\end{lstlisting}

\section{Probe Configuration}
\label{app:probe_config}

\subsection{Label Order}

For the label-order probe, we compare the natural ordinal label space with a reversed label space. In the natural setting, label 1 denotes the most negative class and label 5 denotes the most positive class. In the reversed setting, both the label descriptions and demonstration labels are reversed, so label 1 denotes the most positive class and label 5 denotes the most negative class. After parsing the model output, we map the predicted prompt position back to the original class index: natural order uses \texttt{parsed - 1}, whereas reversed order uses \texttt{M - parsed}, where \texttt{M} is the number of classes.

Natural label order:
\begin{lstlisting}[frame=single,basicstyle=\ttfamily\footnotesize,breaklines=true,breakautoindent=false,breakindent=0pt]
... assign a label from [1: very negative, 2: negative, 3: neutral, 4: positive, 5: very positive].
...
\end{lstlisting}

Reversed label order:
\begin{lstlisting}[frame=single,basicstyle=\ttfamily\footnotesize,breaklines=true,breakautoindent=false,breakindent=0pt]
... assign a label from [1: very positive, 2: positive, 3: neutral, 4: negative, 5: very negative].
...
\end{lstlisting}

\subsection{Demonstration Order}

For the demonstration-order probe, the task header, label semantics, query, and demonstration set are held fixed; only the order of the demonstration blocks is changed. We evaluate the base order, ascending label order, descending label order, and a random permutation. This probe is not applied to the zero-shot setting because no demonstrations are present.

Ascending demo order:
\begin{lstlisting}[frame=single,basicstyle=\ttfamily\footnotesize,breaklines=true,breakautoindent=false,breakindent=0pt]
[header]

Sentence: {demo_text_1_1}
Label: {demo_label_1}

Sentence: {demo_text_1_2}
Label: {demo_label_1}

[... more demos ...]

Sentence: {demo_text_2_1}
Label: {demo_label_2}

Sentence: {demo_text_2_2}
Label: {demo_label_2}

[... more demos ...]

Sentence: {demo_text_K_N}
Label: {demo_label_K}

Sentence: {test_text}
Label:
\end{lstlisting}

\subsection{Demonstration Placement}

For the placement probe, the header and demonstration content are again held fixed, but the demonstrations are positioned differently relative to the query. The baseline places all demonstrations before the query. The \emph{after} condition places the query immediately after the header and appends the demonstrations afterward. The \emph{split} condition places the first half of the demonstrations before the query and the remaining half after the query. Across these variants, the query text and label space remain unchanged; only the relative placement of the demonstration blocks is manipulated.

Before (base):
\begin{lstlisting}[frame=single,basicstyle=\ttfamily\footnotesize,breaklines=true,breakautoindent=false,breakindent=0pt]
[header]

Sentence: {demo_1}
Label: {label_1}

[... more demos ...]

Sentence: {test_text}
Label:

\end{lstlisting}

After:
\begin{lstlisting}[frame=single,basicstyle=\ttfamily\footnotesize,breaklines=true,breakautoindent=false,breakindent=0pt]

[header]

Sentence: {test_text}
Label:

Sentence: {demo_1}
Label: {label_1}

[... more demos ...]

\end{lstlisting}

Split:
\begin{lstlisting}[frame=single,basicstyle=\ttfamily\footnotesize,breaklines=true,breakautoindent=false,breakindent=0pt]
[header]

[... more demos ...]

Sentence: {test_text}
Label:

[... more demos ...]
...
\end{lstlisting}

\section{Phase 1 Configuration}
\label{app:phase1_config}

\subsection{Label Format}

Option format changes the label space in the header and each demonstration's \texttt{Label:} value while preserving the overall \texttt{Sentence:}/\texttt{Label:} block structure.

Numeric (base):
\begin{lstlisting}[frame=single,basicstyle=\ttfamily\footnotesize,breaklines=true,breakautoindent=false,breakindent=0pt]
[header]

Sentence: {demo_text}
Label: 3

[... more demos ...]

Sentence: {test_text}
Label:
\end{lstlisting}

Letter: the label space runs from A (very negative) to E (very positive), and the neutral demonstration uses \texttt{Label: C}.

Natural language: the label space contains the class descriptions directly, and the neutral demonstration uses \texttt{Label: neutral}.

Neutral identifiers: the label space runs from \texttt{Option\_1} (very negative) to \texttt{Option\_5} (very positive), and the neutral demonstration uses \texttt{Label: Option\_3}.

\subsection{Scale Cardinality}

The cardinality factor is evaluated on SST-5 and Yelp-5 by merging the original five classes before prompt construction. The 5-way baseline retains the original labels. In the 3-way condition, the two negative classes are merged, the neutral class is retained, and the two positive classes are merged. In the 2-way condition, the negative and positive classes are each merged and neutral examples are removed. Demonstration labels are remapped to the resulting class space.

3-way (numeric):
\begin{lstlisting}[frame=single,basicstyle=\ttfamily\footnotesize,breaklines=true,breakautoindent=false,breakindent=0pt]
Given the sentence, assign a label from [1: negative, 2: neutral, 3: positive].
\end{lstlisting}

2-way (numeric):
\begin{lstlisting}[frame=single,basicstyle=\ttfamily\footnotesize,breaklines=true,breakautoindent=false,breakindent=0pt]
Given the sentence, assign a label from [1: negative, 2: positive].
\end{lstlisting}

\subsection{Prompt Clarity}

Prompt clarity changes only the task header; the demonstration and query formatting follow the selected option format. The minimal condition provides only a short task instruction, the explicit baseline additionally specifies the label space and output constraint, and the ordinal-explicit condition states the direction of the ordered scale. Under reversed label order, the ordinal-explicit direction is reversed accordingly.

Minimal:
\begin{lstlisting}[frame=single,basicstyle=\ttfamily\footnotesize,breaklines=true,breakautoindent=false,breakindent=0pt]
Classify the sentiment:

Sentence: {demo_text}
Label: 3

[... more demos ...]

Sentence: {test_text}
Label:
\end{lstlisting}

Ordinal-explicit (natural numeric order):
\begin{lstlisting}[frame=single,basicstyle=\ttfamily\footnotesize,breaklines=true,breakautoindent=false,breakindent=0pt]
Classify the sentiment into one of the following ordered categories, where 1 is most negative and 5 is most positive:
Given the sentence, assign a label from [1: very negative, 2: negative, 3: neutral, 4: positive, 5: very positive].
Return label only without any other text.

Sentence: {demo_text}
Label: 3

[... more demos ...]

Sentence: {test_text}
Label:
\end{lstlisting}

\subsection{Separator}

The separator factor changes the punctuation or whitespace between each descriptor (\texttt{Sentence} or \texttt{Label}) and its value. The same separator is used throughout the demonstrations and query, while the header and the connector between lines remain unchanged. The baseline uses a colon followed by a space; the alternatives use a single space or a tab.

Space separator:
\begin{lstlisting}[frame=single,basicstyle=\ttfamily\footnotesize,breaklines=true,breakautoindent=false,breakindent=0pt]
[header]

Sentence {demo_text}
Label 3

[... more demos ...]

Sentence {test_text}
Label
\end{lstlisting}

Tab separator:
\begin{lstlisting}[frame=single,basicstyle=\ttfamily\footnotesize,breaklines=true,breakautoindent=false,breakindent=0pt,tabsize=4]
[header]

Sentence	{demo_text}
Label	3

[... more demos ...]

Sentence	{test_text}
Label
\end{lstlisting}

\subsection{Connector}

The connector factor changes the whitespace joining the \texttt{Sentence} field and the \texttt{Label} field within each demonstration and query block. The baseline places the fields on separate lines. The alternatives place both fields on one line separated by a space or place the label on a new, tab-indented line. The header and separator remain fixed.

Space connector:
\begin{lstlisting}[frame=single,basicstyle=\ttfamily\footnotesize,breaklines=true,breakautoindent=false,breakindent=0pt]
[header]

Sentence: {demo_text} Label: 3

[... more demos ...]

Sentence: {test_text} Label:
\end{lstlisting}

Newline-tab connector:
\begin{lstlisting}[frame=single,basicstyle=\ttfamily\footnotesize,breaklines=true,breakautoindent=false,breakindent=0pt,tabsize=4]
[header]

Sentence: {demo_text}
	Label: 3

[... more demos ...]

Sentence: {test_text}
	Label:
\end{lstlisting}

\subsection{Mood}

The mood factor changes the grammatical form of the task instruction while holding the label space, output constraint, demonstrations, and query fixed. The imperative baseline directly instructs the model to perform the task; the interrogative condition asks the model for a label; and the indicative condition describes the model as performing the task.

Interrogative:
\begin{lstlisting}[frame=single,basicstyle=\ttfamily\footnotesize,breaklines=true,breakautoindent=false,breakindent=0pt]
What is the sentiment of the following sentence?
Which label from [1: very negative, 2: negative, 3: neutral, 4: positive, 5: very positive] do you assign?
Return label only without any other text.
\end{lstlisting}

Indicative:
\begin{lstlisting}[frame=single,basicstyle=\ttfamily\footnotesize,breaklines=true,breakautoindent=false,breakindent=0pt]
You are performing Sentiment Classification task.
Given the sentence, you assign a label from [1: very negative, 2: negative, 3: neutral, 4: positive, 5: very positive].
Return label only without any other text.
\end{lstlisting}

\section{Phase 2 Configuration}
\label{app:phase2_config}

\subsection{Inference Paradigm}

Pairwise. In the pairwise setting, each test instance is scored by comparing the test sentence to every demonstration: for each of the D demos the model is asked twice—once with the test as Passage A and the demo as Passage B (forward), and once with the slots swapped (backward)—which passage is more positive (or more negative under the reversed probe). Each answer is parsed as Passage A or B; forward and backward choices are turned into a consensus score \(F \in \{+1,0,-1\}\), where \(+1\) means the test is more positive than that demo (\(-1\) the opposite; \(0\) for ties or inconsistent answers), and under the reversed probe \(F\) is flipped so it stays in the “more positive” frame. That yields D pairs \((F_i, y_i)\) with demo label \(y_i \in \{1,\ldots,M\}\). The default aggregation method is weighted sum in the form \(S = \sum_i F_i y_i\), which is then mapped onto an equal-width partition of \([-S_{\max}, S_{\max}]\) (where \(S_{\max} = \sum_i y_i\)) into \(M\) bins to produce the final 1-based label—so for 5-class × 5-shot this is 50 LLM calls per instance, then one offline aggregation step.

\begin{lstlisting}[frame=single,basicstyle=\ttfamily\footnotesize,breaklines=true,breakautoindent=false,breakindent=0pt]
Please perform Sentiment Classification task.
Given two Passages, compare their sentiments with labels from ['very negative', 'negative', 'neutral', 'positive', 'very positive'].
Passage A: {test}
Passage B: {demo}
Which Passage is more positive in terms of its sentiment?
Output Passage A or Passage B:
\end{lstlisting}

Listwise. Demonstrations are split into \texttt{k\_shot} class-balanced groups, with one unlabeled demonstration per sentiment class in each group. The model sees each group ordered from most negative to most positive under the natural probe, or in the opposite direction under the reversed probe, and returns the insertion position of the test sentence in the range \(1,\ldots,M\). The \texttt{k\_shot} positions are aggregated by the mean, and then rounded and clipped to \([1,M]\). Under the reversed probe, the aggregated position is mapped back to the natural class index.

\begin{lstlisting}[frame=single,basicstyle=\ttfamily\footnotesize,breaklines=true,breakautoindent=false,breakindent=0pt]
Please perform Sentiment Classification task.
Below are sentences ordered by sentiment from most negative to most positive.
Insert the new sentence into this ordered scale and return only its position (1-5).

Sentence: {class0}
...
Sentence: {class4}

Sentence: {test}
Position:
\end{lstlisting}

Listwise-compare. The same class-balanced groups are used, but each group is presented as labeled reference passages, with one reference per class. For every group, the model is queried twice: once with the natural numeric labels, where 1 is most negative and \(M\) is most positive, and once with the reversed labels. The model selects the reference whose sentiment is closest to the test passage. The \(\texttt{k\_shot}\) positions are aggregated separately by majority vote for each label order. The reversed-order aggregate is then mapped back to the natural class space, and P1 is computed between the two aggregated predictions.

\begin{lstlisting}[frame=single,basicstyle=\ttfamily\footnotesize,breaklines=true,breakautoindent=false,breakindent=0pt]
Please perform a Sentiment Classification task. Given the sentence, assign a label from [1: very negative, 2: negative, 3: neutral, 4: positive, 5: very positive].

Given 5 reference passages ordered by sentiment from very negative to very positive:

Sentence: {demo}
Label: 1
...

New passage: {test}

Which reference (1-5) is closest in sentiment to the new passage?
Answer with ONLY a single number (1, 2, 3, 4, or 5). No explanation.
\end{lstlisting}

\subsection{Pairwise Aggregation}

For each test instance \(x\), pairwise inference compares it with every demonstration \(x_i\), whose class label is \(y_i \in \{1,\ldots,M\}\). Forward and backward comparisons are combined into a consensus outcome
\[
F(x,x_i) \in \{+1,0,-1\},
\]
where \(+1\) indicates that the test is more positive, \(-1\) indicates that the demonstration is more positive, and \(0\) represents a tie or inconsistent comparison. The five aggregation methods operate offline on the resulting \((F(x,x_i),y_i)\) pairs and therefore require no additional prompts or model calls beyond the shared pairwise comparisons.

\medskip\noindent\textbf{Majority vote.}\par
Each demonstration produces a discrete class vote that shifts its label by one position in the direction indicated by the comparison:
\[
v_i=
\begin{cases}
\min(y_i+1,M), & F(x,x_i)=+1,\\
y_i, & F(x,x_i)=0,\\
\max(y_i-1,1), & F(x,x_i)=-1.
\end{cases}
\]
The prediction is \(\hat{y}=\operatorname{mode}(v_1,\ldots,v_N)\).
For example, beating demonstrations with labels 1 and 3 and losing to a demonstration with label 5 produces votes \(\{2,4,4\}\), yielding class 4.

\medskip\noindent\textbf{Mean score.}\par
The mean-score method uses the same shifted votes but averages them rather than taking their mode:
\[
\hat{y}
=\operatorname{clip}\!\left(
\operatorname{round}\!\left(\frac{1}{N}\sum_{i=1}^{N}v_i\right),
1,M\right).
\]
Thus, votes \(\{2,4,4\}\) have mean \(3.33\) and produce class 3.

\medskip\noindent\textbf{Weighted sum.}\par
The default pairwise aggregator weights each comparison by the demonstration label:
\[
S(x)=\sum_{i=1}^{N}F(x,x_i)y_i,
\qquad
S_{\max}=\sum_{i=1}^{N}y_i.
\]
The score lies in \([-S_{\max},S_{\max}]\), which is divided into \(M\) equal-width bins:
\[
\hat{y}
=\operatorname{clip}\!\left(
\left\lfloor
\frac{S(x)+S_{\max}}{2S_{\max}/M}
\right\rfloor+1,
1,M\right).
\]
For SST-5 with five demonstrations per class, \(S_{\max}=75\), giving bins of width 30. This weighting makes a comparison with a label-5 demonstration contribute five times as much as a comparison with a label-1 demonstration.

\medskip\noindent\textbf{Copeland score.}\par
Copeland aggregation ignores demonstration labels and counts only net wins:
\[
\operatorname{net}(x)=\sum_{i=1}^{N}F(x,x_i) \in [-N,N],
\]
followed by equal-width binning over that interval:
\[
\hat{y}
=\operatorname{clip}\!\left(
\left\lfloor
\frac{\operatorname{net}(x)+N}{2N/M}
\right\rfloor+1,
1,M\right).
\]
With \(N=25\) and \(M=5\), the bins have width 10, and wins against weak and strong demonstrations contribute equally.

\medskip\noindent\textbf{Bradley--Terry.}\par
This method treats the \(N\) demonstrations and the test instance as participants with latent positive strengths \(\theta\), using
\[
P(A \succ B)=\frac{\theta_A}{\theta_A+\theta_B}.
\]
Pairwise outcomes define a win matrix, with ties assigning half a win to each participant. Strengths are iteratively updated by
\[
\theta_i
\leftarrow
\frac{w_i}{\displaystyle\sum_{j\in\operatorname{opp}(i)}
1/(\theta_i+\theta_j)},
\]
and normalized to have mean one after each iteration. We use \texttt{max\_iter=100} and \texttt{tol=1e-6}. The final prediction is the class whose demonstrations have mean strength closest to the test strength:
\[
\hat{y}=\arg\min_c
\left|\bar{\theta}_c-\theta_{\mathrm{test}}\right|,
\]
where \(\bar{\theta}_c\) is the mean strength of demonstrations labeled \(c\).

\subsection{Debiasing Strategies}

All four debiasing methods start from the same pointwise ICL baseline, with the model, demonstrations, test instances, and seeds held fixed. Label-order and demonstration-order averaging reuse existing hard predictions and require no new prompts. PriDe retains the baseline prompt but requests output log probabilities, whereas Contextual Calibration is the only method that changes the query text.

\medskip\noindent\textbf{Label-order averaging.}\par
This method combines the natural- and reversed-order predictions from Probe~1. After both predictions are mapped back to the common semantic class space, the final prediction is
\[
\hat{y}
=\operatorname{round}\!\left(
\frac{y_{\mathrm{nat}}+y_{\mathrm{rev}}}{2}
\right).
\]
For example, \(y_{\mathrm{nat}}=3\) and \(y_{\mathrm{rev}}=4\) produce \(\hat{y}=4\). Because the two label-order conditions are merged into one output, this method consumes the label-order axis. We evaluate the resulting averaged prediction as a single output and therefore do not report residual PFRs for this method. It requires no additional LLM calls.

\medskip\noindent\textbf{Demonstration-order averaging.}\par
This method analogously averages the class-ascending and class-descending predictions from Probe~2:
\[
\hat{y}
=\operatorname{round}\!\left(
\frac{y_{\mathrm{asc}}+y_{\mathrm{desc}}}{2}
\right).
\]
It consumes the corresponding demonstration-order contrast. We evaluate the resulting averaged prediction as a single output and therefore do not report residual PFRs for this method. It requires no additional prompts or LLM calls.

\medskip\noindent\textbf{PriDe.}\par
PriDe~\citep{zheng2024large} reruns every test instance under both natural and reversed label orders while recording the top-$k$ output log probabilities, requiring approximately 400 model calls per dataset and seed. Using 10 calibration instances evaluated under both label-order permutations, it estimates the position prior for label position $k$ as
\[
\operatorname{prior}(k)
= \underset{x,\,\pi}{\operatorname{mean}}\, P(k \mid x,\pi),
\]
where $x$ ranges over calibration instances and $\pi$ ranges over the natural and reversed permutations. Averaging across permutations reduces the contribution of instance content and isolates the model's preference for particular label positions.

For each test instance, PriDe divides the raw probability assigned to each position by its estimated prior and renormalizes the resulting scores:
\[
P_{\mathrm{corr}}(k)
\propto \frac{P_{\mathrm{raw}}(k)}{\operatorname{prior}(k)},
\qquad
\hat{y}=\arg\max_k P_{\mathrm{corr}}(k).
\]
This correction is applied separately to the natural- and reversed-order runs, after which reversed predictions are mapped back to the natural class space and residual PFR is computed between the two corrected predictions. The calibration set contains 10 instances, and the retained log-probability distribution contains the top 20 tokens.

\medskip\noindent\textbf{Contextual Calibration.}\par
Contextual Calibration~\citep{zhao2021calibrate} estimates prompt-induced label bias using a content-free query. For each label order, the task header, label space, and demonstrations are kept unchanged, while the test sentence is replaced with \texttt{N/A}:

\begin{lstlisting}[frame=single,basicstyle=\ttfamily\footnotesize,breaklines=true,breakautoindent=false,breakindent=0pt]
[header and label space]

[demonstrations unchanged]

Sentence: N/A
Label:
\end{lstlisting}

This requires one null-prompt call per label order; the test-instance log probabilities are reused from the same natural- and reversed-order runs used by PriDe. For each order, the null distribution \(P_{\mathrm{null}}(k)\) represents the probability assigned to label position \(k\) in the absence of meaningful test content. Each test distribution is corrected as
\[
P_{\mathrm{corr}}(k)
\propto
\frac{P_{\mathrm{raw}}(k)}{\max\!\left(P_{\mathrm{null}}(k),10^{-10}\right)}.
\]
The corrected prediction is \(\hat{y}=\arg\max_k P_{\mathrm{corr}}(k)\).
The probability floor prevents numerical instability when a null probability is close to zero. Correction is applied separately to each label order, after which reversed predictions are mapped back to the natural class space and residual PFR is computed from the corrected predictions.

\FloatBarrier
\input{appendix/full_phase_tables}
\FloatBarrier

\section{Priority Configuration}
\label{app:priority_config}

Table~\ref{tab:priority_config} lists the fixed Phase~1 factor combinations constructed for each optimization objective. Each combination selects one level per factor and is evaluated without further tuning during the cross-model transfer analysis.

\begin{table*}[t]
\centering
\small
\resizebox{\textwidth}{!}{%
\begin{tabular}{clllllll}
\toprule
ID & Prioritization & F1 & F6 & F13 & Separator & Connector & Mood \\
\midrule
\textbf{A} & Accuracy $\uparrow$ & \texttt{neutral\_id} & 0 & \texttt{ordinal\_explicit} & \emph{base} & \emph{base} & \texttt{indicative} \\
\textbf{P1} & Probe 1 PFR $\downarrow$ & \emph{base} (\texttt{numeric}) & 3 & \emph{base} (\texttt{explicit}) & \emph{base} & \emph{base} & \texttt{indicative} \\
\textbf{P2a} & Probe 2 \texttt{random\_vs\_ascending} $\downarrow$ & \texttt{natural\_language} & 1 & \texttt{minimal} & \emph{base} & \emph{base} & \texttt{indicative} \\
\textbf{P2b} & Probe 2 \texttt{random\_vs\_descending} $\downarrow$ & \texttt{neutral\_id} & 1 & \emph{base} (\texttt{explicit}) & \emph{base} & \emph{base} & \texttt{indicative} \\
\textbf{P3a} & Probe 3 \texttt{before\_vs\_after} $\downarrow$ & \texttt{neutral\_id} & 3 & \texttt{minimal} & \texttt{tab} & \texttt{space} & \texttt{interrogative} \\
\textbf{P3b} & Probe 3 \texttt{before\_vs\_split} $\downarrow$ & \texttt{neutral\_id} & 1 & \texttt{minimal} & \emph{base} & \texttt{newline\_tab} & \emph{base} \\
\bottomrule
\end{tabular}
}
\caption{Objective-specific Phase~1 factor combinations used in the priority and cross-model transfer analyses.}
\label{tab:priority_config}
\end{table*}

\input{appendix/aggregation_table}

%% file: appendix/full_phase_tables.tex
\section{Full Phase 1 and Phase 2 Results}
\label{app:full_phase_results}

Tables~\ref{tab:phase1_residuals_full_a}--\ref{tab:phase2_pipeline_full} report the complete means and standard deviations underlying the main-paper residual tables. Table~\ref{tab:phase2_joint_full} gives the absolute joint-factor results used to compute the main-paper cross-model residuals. Table~\ref{tab:phase2_pipeline_full} gives the full inference-paradigm, pairwise-aggregation, and debiasing results summarized in the main paper.

\providecolor{phasePositive}{HTML}{C65F5F}
\providecolor{phaseNegative}{HTML}{4C78A8}
\providecolor{priorityPositive}{HTML}{C65F5F}
\providecolor{priorityNegative}{HTML}{4C78A8}

\newcommand{\phasefullcell}[5]{\colorbox{#1!#2}{\textcolor{#3}{\makebox[3.65em][c]{\shortstack{$#4$\\$(\pm #5)$}}}}}
\newcommand{\phasefullna}{\colorbox{gray!20}{\makebox[3.65em][c]{\shortstack{--\\\phantom{$(\pm 0.00)$}}}}}
\newcommand{\priorityfullcell}[5]{\colorbox{#1!#2}{\textcolor{#3}{\makebox[3.70em][c]{\shortstack{$#4$\\$(\pm #5)$}}}}}
\newcommand{\priorityfullna}{\colorbox{gray!20}{\makebox[3.70em][c]{\shortstack{--\\\phantom{$(\pm 0.000)$}}}}}

\begin{table*}[t]
\centering
\small
\setlength{\tabcolsep}{0.8pt}
\renewcommand{\arraystretch}{0.88}
\begin{tabular}{l@{\hspace{4pt}}lccccccccc}
\toprule
& & \multicolumn{4}{c}{Performance residuals} & \multicolumn{5}{c}{Position-sensitivity residuals} \\
Factor & Level & Acc. & F1 & $\rho$ & MAE & P1 & P2a & P2b & P3a & P3b \\
\midrule
Label format & Letter & \phasefullcell{phasePositive}{11}{black}{+2.20}{2.72} & \phasefullcell{phasePositive}{11}{black}{+2.71}{4.08} & \phasefullcell{phasePositive}{12}{black}{+3.06}{3.86} & \phasefullcell{phaseNegative}{10}{black}{-1.53}{3.72} & \phasefullcell{phasePositive}{12}{black}{+3.27}{5.15} & \phasefullcell{phasePositive}{9}{black}{+1.18}{7.38} & \phasefullcell{phasePositive}{10}{black}{+1.87}{6.52} & \phasefullcell{phasePositive}{13}{black}{+4.04}{15.33} & \phasefullcell{phasePositive}{10}{black}{+1.27}{5.32} \\
 & Natural language & \phasefullcell{phasePositive}{10}{black}{+1.27}{5.11} & \phasefullcell{phasePositive}{8}{black}{+0.35}{6.53} & \phasefullcell{phasePositive}{13}{black}{+4.01}{3.06} & \phasefullcell{phaseNegative}{10}{black}{-1.87}{5.93} & \phasefullcell{phasePositive}{80}{white}{+59.54}{27.81} & \phasefullcell{phaseNegative}{24}{black}{-13.18}{14.10} & \phasefullcell{phaseNegative}{10}{black}{-1.96}{4.70} & \phasefullcell{phasePositive}{32}{black}{+19.51}{24.07} & \phasefullcell{phasePositive}{13}{black}{+4.44}{11.41} \\
 & Neutral ID & \phasefullcell{phasePositive}{16}{black}{+6.53}{2.79} & \phasefullcell{phasePositive}{18}{black}{+8.10}{3.95} & \phasefullcell{phasePositive}{15}{black}{+6.12}{4.93} & \phasefullcell{phaseNegative}{17}{black}{-7.33}{3.41} & \phasefullcell{phasePositive}{9}{black}{+0.50}{2.02} & \phasefullcell{phaseNegative}{17}{black}{-7.32}{11.69} & \phasefullcell{phaseNegative}{17}{black}{-7.26}{6.94} & \phasefullcell{phaseNegative}{21}{black}{-10.87}{14.00} & \phasefullcell{phaseNegative}{24}{black}{-12.95}{9.25} \\
\midrule
Model & Qwen3-8B Base & \phasefullcell{phasePositive}{17}{black}{+7.17}{2.06} & \phasefullcell{phasePositive}{19}{black}{+9.09}{4.65} & \phasefullcell{phasePositive}{16}{black}{+6.44}{5.44} & \phasefullcell{phaseNegative}{18}{black}{-8.33}{3.33} & \phasefullcell{phaseNegative}{9}{black}{-0.73}{3.90} & \phasefullcell{phaseNegative}{12}{black}{-3.42}{7.82} & \phasefullcell{phasePositive}{23}{black}{+12.47}{9.96} & \phasefullcell{phasePositive}{32}{black}{+20.13}{10.57} & \phasefullcell{phasePositive}{40}{black}{+26.19}{9.52} \\
& Qwen3-8B Thinking & \phasefullcell{phasePositive}{12}{black}{+3.05}{1.87} & \phasefullcell{phasePositive}{14}{black}{+4.72}{2.96} & \phasefullcell{phasePositive}{11}{black}{+2.73}{5.33} & \phasefullcell{phaseNegative}{10}{black}{-1.66}{3.71} & \phasefullcell{phasePositive}{19}{black}{+9.38}{4.31} & \phasefullcell{phaseNegative}{19}{black}{-8.77}{13.15} & \phasefullcell{phaseNegative}{11}{black}{-2.11}{12.78} & \phasefullcell{phaseNegative}{33}{black}{-21.01}{12.32} & \phasefullcell{phaseNegative}{30}{black}{-18.32}{13.72} \\
& Qwen3-4B & \phasefullcell{phasePositive}{13}{black}{+4.17}{5.75} & \phasefullcell{phasePositive}{16}{black}{+6.30}{7.14} & \phasefullcell{phasePositive}{12}{black}{+3.45}{3.06} & \phasefullcell{phaseNegative}{14}{black}{-4.87}{6.37} & \phasefullcell{phaseNegative}{13}{black}{-4.07}{4.44} & \phasefullcell{phaseNegative}{24}{black}{-13.25}{12.15} & \phasefullcell{phaseNegative}{20}{black}{-10.33}{8.35} & \phasefullcell{phasePositive}{17}{black}{+7.48}{9.91} & \phasefullcell{phasePositive}{14}{black}{+5.34}{15.36} \\
& Gemma3-12B & \phasefullcell{phasePositive}{22}{black}{+11.46}{3.24} & \phasefullcell{phasePositive}{25}{black}{+13.79}{5.22} & \phasefullcell{phasePositive}{19}{black}{+8.73}{3.90} & \phasefullcell{phaseNegative}{24}{black}{-13.16}{5.08} & \phasefullcell{phaseNegative}{12}{black}{-2.98}{7.91} & \phasefullcell{phaseNegative}{25}{black}{-13.98}{14.62} & \phasefullcell{phaseNegative}{18}{black}{-7.93}{9.72} & \phasefullcell{phaseNegative}{26}{black}{-15.13}{20.28} & \phasefullcell{phaseNegative}{26}{black}{-14.96}{17.68} \\
& Ministral-3-8B & \phasefullcell{phasePositive}{9}{black}{+0.93}{2.12} & \phasefullcell{phasePositive}{9}{black}{+0.49}{1.37} & \phasefullcell{phaseNegative}{8}{black}{-0.22}{0.49} & \phasefullcell{phaseNegative}{9}{black}{-0.90}{2.15} & \phasefullcell{phasePositive}{9}{black}{+0.84}{1.50} & \phasefullcell{phasePositive}{9}{black}{+0.48}{0.37} & \phasefullcell{phasePositive}{9}{black}{+0.94}{1.91} & \phasefullcell{phasePositive}{12}{black}{+3.71}{7.47} & \phasefullcell{phaseNegative}{8}{black}{-0.16}{0.86} \\
& Gemma4-12B & \phasefullcell{phasePositive}{24}{black}{+13.40}{2.46} & \phasefullcell{phasePositive}{27}{black}{+15.49}{5.24} & \phasefullcell{phasePositive}{22}{black}{+11.92}{6.04} & \phasefullcell{phaseNegative}{26}{black}{-14.80}{4.27} & \phasefullcell{phaseNegative}{16}{black}{-7.00}{6.85} & \phasefullcell{phaseNegative}{33}{black}{-20.75}{16.52} & \phasefullcell{phaseNegative}{29}{black}{-17.60}{15.62} & \phasefullcell{phaseNegative}{30}{black}{-17.81}{21.88} & \phasefullcell{phaseNegative}{52}{black}{-36.50}{13.36} \\
& Granite-4-8B & \phasefullcell{phasePositive}{19}{black}{+9.17}{4.72} & \phasefullcell{phasePositive}{21}{black}{+10.97}{5.97} & \phasefullcell{phasePositive}{17}{black}{+7.83}{3.78} & \phasefullcell{phaseNegative}{21}{black}{-10.90}{6.56} & \phasefullcell{phaseNegative}{12}{black}{-3.33}{4.30} & \phasefullcell{phaseNegative}{29}{black}{-17.08}{12.96} & \phasefullcell{phaseNegative}{20}{black}{-10.29}{8.36} & \phasefullcell{phaseNegative}{10}{black}{-1.72}{7.84} & \phasefullcell{phasePositive}{21}{black}{+11.01}{17.71} \\
\midrule
Scale cardinality & 3-way & \phasefullcell{phasePositive}{38}{black}{+25.00}{6.17} & \phasefullcell{phasePositive}{37}{black}{+23.84}{6.20} & \phasefullcell{phasePositive}{11}{black}{+2.53}{4.30} & \phasefullcell{phaseNegative}{44}{black}{-29.42}{7.08} & \phasefullcell{phaseNegative}{28}{black}{-16.33}{0.34} & \phasefullcell{phaseNegative}{47}{black}{-32.37}{9.46} & \phasefullcell{phaseNegative}{40}{black}{-26.41}{3.74} & \phasefullcell{phaseNegative}{29}{black}{-17.22}{0.45} & \phasefullcell{phaseNegative}{39}{black}{-25.31}{4.51} \\
 & 2-way & \phasefullcell{phasePositive}{61}{white}{+43.85}{5.57} & \phasefullcell{phasePositive}{65}{white}{+47.47}{5.39} & \phasefullcell{phasePositive}{11}{black}{+2.47}{8.85} & \phasefullcell{phaseNegative}{68}{white}{-49.34}{6.90} & \phasefullcell{phaseNegative}{29}{black}{-17.34}{0.32} & \phasefullcell{phaseNegative}{65}{white}{-47.41}{8.21} & \phasefullcell{phaseNegative}{51}{black}{-35.57}{3.94} & \phasefullcell{phaseNegative}{48}{black}{-33.07}{3.85} & \phasefullcell{phaseNegative}{67}{white}{-48.91}{4.42} \\
\bottomrule
\end{tabular}
\caption{Full Phase~1 residuals (mean $\pm$ SD): label format, model, and scale cardinality.}
\label{tab:phase1_residuals_full_a}
\end{table*}

\begin{table*}[t]
\centering
\small
\setlength{\tabcolsep}{0.8pt}
\renewcommand{\arraystretch}{0.88}
\begin{tabular}{l@{\hspace{4pt}}lccccccccc}
\toprule
& & \multicolumn{4}{c}{Performance residuals} & \multicolumn{5}{c}{Position-sensitivity residuals} \\
Factor & Level & Acc. & F1 & $\rho$ & MAE & P1 & P2a & P2b & P3a & P3b \\
\midrule
Demo count & 0-shot & \phasefullcell{phasePositive}{17}{black}{+7.13}{3.49} & \phasefullcell{phasePositive}{18}{black}{+8.01}{6.04} & \phasefullcell{phasePositive}{16}{black}{+6.45}{4.82} & \phasefullcell{phaseNegative}{17}{black}{-7.37}{4.74} & \phasefullcell{phasePositive}{16}{black}{+6.87}{4.11} & \phasefullna & \phasefullna & \phasefullna & \phasefullna \\
 & 1-shot & \phasefullcell{phasePositive}{13}{black}{+4.40}{3.63} & \phasefullcell{phasePositive}{13}{black}{+4.00}{5.69} & \phasefullcell{phasePositive}{15}{black}{+5.79}{4.63} & \phasefullcell{phaseNegative}{14}{black}{-4.83}{3.60} & \phasefullcell{phasePositive}{12}{black}{+3.43}{6.64} & \phasefullcell{phaseNegative}{28}{black}{-16.78}{14.23} & \phasefullcell{phaseNegative}{17}{black}{-7.33}{3.45} & \phasefullcell{phaseNegative}{16}{black}{-6.25}{11.80} & \phasefullcell{phaseNegative}{32}{black}{-20.16}{13.06} \\
 & 3-shot & \phasefullcell{phasePositive}{9}{black}{+0.77}{2.36} & \phasefullcell{phasePositive}{9}{black}{+0.80}{3.41} & \phasefullcell{phasePositive}{10}{black}{+1.78}{1.34} & \phasefullcell{phaseNegative}{10}{black}{-1.53}{2.10} & \phasefullcell{phaseNegative}{8}{black}{-0.13}{2.77} & \phasefullcell{phaseNegative}{19}{black}{-8.72}{10.49} & \phasefullcell{phaseNegative}{15}{black}{-6.20}{5.36} & \phasefullcell{phaseNegative}{26}{black}{-14.62}{12.19} & \phasefullcell{phaseNegative}{23}{black}{-12.09}{12.51} \\
\midrule
Prompt clarity & Minimal & \phasefullcell{phaseNegative}{12}{black}{-3.68}{17.52} & \phasefullcell{phaseNegative}{18}{black}{-8.06}{17.81} & \phasefullcell{phaseNegative}{39}{black}{-25.57}{19.42} & \phasefullcell{phasePositive}{29}{black}{+17.50}{32.40} & \phasefullcell{phasePositive}{67}{white}{+48.84}{27.54} & \phasefullcell{phaseNegative}{22}{black}{-11.89}{25.13} & \phasefullcell{phasePositive}{18}{black}{+8.43}{22.60} & \phasefullcell{phaseNegative}{24}{black}{-13.50}{22.02} & \phasefullcell{phaseNegative}{23}{black}{-12.21}{14.15} \\
 & Ordinal-explicit & \phasefullcell{phasePositive}{11}{black}{+2.77}{3.40} & \phasefullcell{phasePositive}{10}{black}{+1.60}{4.55} & \phasefullcell{phaseNegative}{11}{black}{-2.33}{8.14} & \phasefullcell{phaseNegative}{13}{black}{-4.10}{4.30} & \phasefullcell{phasePositive}{14}{black}{+4.97}{1.83} & \phasefullcell{phasePositive}{12}{black}{+3.48}{12.35} & \phasefullcell{phasePositive}{18}{black}{+8.54}{11.62} & \phasefullcell{phasePositive}{12}{black}{+3.38}{18.35} & \phasefullcell{phaseNegative}{16}{black}{-6.89}{1.69} \\
\midrule
Separator & Space & \phasefullcell{phaseNegative}{15}{black}{-5.50}{6.86} & \phasefullcell{phaseNegative}{16}{black}{-6.37}{8.03} & \phasefullcell{phaseNegative}{14}{black}{-5.07}{8.33} & \phasefullcell{phasePositive}{18}{black}{+7.95}{11.61} & \phasefullcell{phasePositive}{8}{black}{+0.22}{1.63} & \phasefullcell{phasePositive}{15}{black}{+5.38}{10.47} & \phasefullcell{phasePositive}{22}{black}{+11.20}{12.80} & \phasefullcell{phasePositive}{10}{black}{+1.72}{10.33} & \phasefullcell{phasePositive}{9}{black}{+0.93}{22.27} \\
 & Tab & \phasefullcell{phaseNegative}{13}{black}{-4.37}{8.84} & \phasefullcell{phaseNegative}{15}{black}{-5.81}{12.29} & \phasefullcell{phaseNegative}{16}{black}{-6.34}{16.49} & \phasefullcell{phasePositive}{22}{black}{+11.43}{22.99} & \phasefullcell{phasePositive}{8}{black}{+0.33}{5.99} & \phasefullcell{phasePositive}{20}{black}{+9.56}{18.80} & \phasefullcell{phasePositive}{27}{black}{+15.57}{17.49} & \phasefullcell{phaseNegative}{10}{black}{-1.59}{13.85} & \phasefullcell{phasePositive}{10}{black}{+1.55}{23.66} \\
\midrule
Connector & Space & \phasefullcell{phaseNegative}{18}{black}{-7.86}{8.53} & \phasefullcell{phaseNegative}{18}{black}{-8.17}{8.10} & \phasefullcell{phaseNegative}{19}{black}{-9.28}{8.27} & \phasefullcell{phasePositive}{22}{black}{+11.50}{10.25} & \phasefullcell{phasePositive}{9}{black}{+0.61}{2.92} & \phasefullcell{phasePositive}{16}{black}{+6.62}{7.51} & \phasefullcell{phasePositive}{27}{black}{+15.56}{9.57} & \phasefullcell{phaseNegative}{16}{black}{-6.23}{11.80} & \phasefullcell{phaseNegative}{10}{black}{-2.06}{12.43} \\
 & Newline+tab & \phasefullcell{phaseNegative}{12}{black}{-3.23}{2.78} & \phasefullcell{phaseNegative}{13}{black}{-3.91}{2.80} & \phasefullcell{phaseNegative}{12}{black}{-3.43}{3.22} & \phasefullcell{phasePositive}{13}{black}{+3.76}{2.61} & \phasefullcell{phasePositive}{10}{black}{+1.68}{2.55} & \phasefullcell{phasePositive}{9}{black}{+0.89}{0.58} & \phasefullcell{phasePositive}{12}{black}{+3.05}{2.26} & \phasefullcell{phaseNegative}{10}{black}{-2.02}{6.60} & \phasefullcell{phaseNegative}{16}{black}{-6.77}{4.28} \\
\midrule
Mood & Interrogative & \phasefullcell{phaseNegative}{12}{black}{-3.57}{4.46} & \phasefullcell{phaseNegative}{14}{black}{-5.37}{3.76} & \phasefullcell{phaseNegative}{14}{black}{-4.67}{5.63} & \phasefullcell{phasePositive}{12}{black}{+3.17}{4.55} & \phasefullcell{phasePositive}{13}{black}{+3.97}{3.78} & \phasefullcell{phasePositive}{9}{black}{+0.65}{2.79} & \phasefullcell{phasePositive}{33}{black}{+20.50}{16.07} & \phasefullcell{phaseNegative}{15}{black}{-6.14}{8.57} & \phasefullcell{phasePositive}{10}{black}{+1.82}{11.22} \\
 & Indicative & \phasefullcell{phasePositive}{10}{black}{+1.97}{1.61} & \phasefullcell{phasePositive}{11}{black}{+2.42}{2.11} & \phasefullcell{phasePositive}{10}{black}{+1.33}{1.76} & \phasefullcell{phaseNegative}{12}{black}{-2.90}{2.59} & \phasefullcell{phaseNegative}{8}{black}{-0.17}{2.08} & \phasefullcell{phaseNegative}{12}{black}{-3.65}{4.06} & \phasefullcell{phaseNegative}{12}{black}{-3.39}{4.07} & \phasefullcell{phasePositive}{14}{black}{+4.59}{3.67} & \phasefullcell{phasePositive}{11}{black}{+2.30}{1.52} \\
\bottomrule
\end{tabular}
\caption{Full Phase~1 residuals (mean $\pm$ SD): demonstration count and prompt-form factors.}
\label{tab:phase1_residuals_full_b}
\end{table*}

\begin{table*}[t]
\centering
\small
\setlength{\tabcolsep}{0.5pt}
\renewcommand{\arraystretch}{0.78}
\setlength{\aboverulesep}{0pt}
\setlength{\belowrulesep}{0pt}
\begin{tabular}{l@{\hspace{4pt}}lccccccccc}
\toprule
& & \multicolumn{4}{c}{Performance} & \multicolumn{5}{c}{Position sensitivity} \\
Model & Config. & Acc. & F1 & $\rho$ & MAE & P1 & P2a & P2b & P3a & P3b \\
\midrule
Ministral-3-8B & baseline & \priorityfullcell{priorityPositive}{11}{black}{0.565}{0.114} & \priorityfullcell{priorityPositive}{10}{black}{0.522}{0.117} & \priorityfullcell{priorityNegative}{9}{black}{0.577}{0.220} & \priorityfullcell{priorityNegative}{11}{black}{0.460}{0.134} & \priorityfullcell{priorityPositive}{11}{black}{0.136}{0.087} & \priorityfullcell{priorityPositive}{10}{black}{0.312}{0.217} & \priorityfullcell{priorityPositive}{11}{black}{0.270}{0.155} & \priorityfullcell{priorityPositive}{21}{black}{0.577}{0.193} & \priorityfullcell{priorityNegative}{9}{black}{0.484}{0.175} \\
 & A & \priorityfullcell{priorityNegative}{32}{black}{0.533}{0.060} & \priorityfullcell{priorityNegative}{40}{black}{0.494}{0.083} & \priorityfullcell{priorityNegative}{29}{black}{0.578}{0.264} & \priorityfullcell{priorityPositive}{32}{black}{0.498}{0.081} & \priorityfullcell{priorityPositive}{66}{white}{0.450}{0.255} & \priorityfullna & \priorityfullna & \priorityfullna & \priorityfullna \\
 & P1 & \priorityfullcell{priorityNegative}{20}{black}{0.550}{0.087} & \priorityfullcell{priorityNegative}{18}{black}{0.527}{0.096} & \priorityfullcell{priorityNegative}{23}{black}{0.567}{0.211} & \priorityfullcell{priorityPositive}{27}{black}{0.485}{0.117} & \priorityfullcell{priorityPositive}{26}{black}{0.171}{0.095} & \priorityfullcell{priorityPositive}{10}{black}{0.214}{0.124} & \priorityfullcell{priorityPositive}{35}{black}{0.269}{0.137} & \priorityfullcell{priorityNegative}{12}{black}{0.408}{0.128} & \priorityfullcell{priorityNegative}{12}{black}{0.383}{0.139} \\
 & P2a & \priorityfullcell{priorityNegative}{44}{black}{0.522}{0.079} & \priorityfullcell{priorityNegative}{48}{black}{0.486}{0.107} & \priorityfullcell{priorityNegative}{44}{black}{0.539}{0.281} & \priorityfullcell{priorityPositive}{50}{black}{0.510}{0.104} & \priorityfullcell{priorityPositive}{11}{black}{0.827}{0.153} & \priorityfullcell{priorityPositive}{43}{black}{0.227}{0.197} & \priorityfullcell{priorityPositive}{29}{black}{0.252}{0.179} & \priorityfullcell{priorityNegative}{22}{black}{0.306}{0.183} & \priorityfullcell{priorityNegative}{12}{black}{0.314}{0.213} \\
 & P2b & \priorityfullcell{priorityNegative}{19}{black}{0.590}{0.090} & \priorityfullcell{priorityNegative}{17}{black}{0.570}{0.102} & \priorityfullcell{priorityNegative}{19}{black}{0.622}{0.230} & \priorityfullcell{priorityPositive}{21}{black}{0.430}{0.102} & \priorityfullcell{priorityNegative}{40}{black}{0.083}{0.067} & \priorityfullcell{priorityPositive}{21}{black}{0.132}{0.125} & \priorityfullcell{priorityPositive}{19}{black}{0.154}{0.089} & \priorityfullcell{priorityPositive}{43}{black}{0.469}{0.267} & \priorityfullcell{priorityPositive}{18}{black}{0.377}{0.236} \\
\midrule
Gemma3-12B & baseline & \priorityfullcell{priorityPositive}{47}{black}{0.670}{0.075} & \priorityfullcell{priorityPositive}{55}{white}{0.655}{0.065} & \priorityfullcell{priorityPositive}{38}{black}{0.667}{0.201} & \priorityfullcell{priorityNegative}{53}{black}{0.337}{0.077} & \priorityfullcell{priorityNegative}{18}{black}{0.098}{0.042} & \priorityfullcell{priorityNegative}{56}{white}{0.168}{0.074} & \priorityfullcell{priorityNegative}{35}{black}{0.181}{0.135} & \priorityfullcell{priorityNegative}{60}{white}{0.389}{0.228} & \priorityfullcell{priorityNegative}{59}{white}{0.336}{0.237} \\
 & A & \priorityfullcell{priorityNegative}{16}{black}{0.581}{0.064} & \priorityfullcell{priorityNegative}{16}{black}{0.562}{0.080} & \priorityfullcell{priorityNegative}{9}{black}{0.637}{0.236} & \priorityfullcell{priorityPositive}{12}{black}{0.438}{0.071} & \priorityfullcell{priorityNegative}{31}{black}{0.211}{0.069} & \priorityfullna & \priorityfullna & \priorityfullna & \priorityfullna \\
 & P1 & \priorityfullcell{priorityPositive}{38}{black}{0.672}{0.092} & \priorityfullcell{priorityPositive}{43}{black}{0.660}{0.076} & \priorityfullcell{priorityPositive}{31}{black}{0.680}{0.184} & \priorityfullcell{priorityNegative}{39}{black}{0.338}{0.099} & \priorityfullcell{priorityPositive}{18}{black}{0.146}{0.061} & \priorityfullcell{priorityNegative}{23}{black}{0.164}{0.079} & \priorityfullcell{priorityNegative}{13}{black}{0.175}{0.080} & \priorityfullcell{priorityPositive}{68}{white}{0.598}{0.158} & \priorityfullcell{priorityPositive}{47}{black}{0.511}{0.150} \\
 & P2a & \priorityfullcell{priorityPositive}{13}{black}{0.644}{0.090} & \priorityfullcell{priorityPositive}{12}{black}{0.613}{0.105} & \priorityfullcell{priorityPositive}{16}{black}{0.669}{0.205} & \priorityfullcell{priorityNegative}{16}{black}{0.363}{0.092} & \priorityfullcell{priorityPositive}{10}{black}{0.824}{0.136} & \priorityfullcell{priorityPositive}{13}{black}{0.139}{0.042} & \priorityfullcell{priorityNegative}{16}{black}{0.168}{0.044} & \priorityfullcell{priorityNegative}{35}{black}{0.267}{0.239} & \priorityfullcell{priorityNegative}{46}{black}{0.213}{0.156} \\
 & P2b & \priorityfullcell{priorityPositive}{23}{black}{0.666}{0.104} & \priorityfullcell{priorityPositive}{24}{black}{0.644}{0.112} & \priorityfullcell{priorityPositive}{17}{black}{0.681}{0.186} & \priorityfullcell{priorityNegative}{25}{black}{0.342}{0.106} & \priorityfullcell{priorityNegative}{22}{black}{0.136}{0.054} & \priorityfullcell{priorityPositive}{16}{black}{0.117}{0.059} & \priorityfullcell{priorityPositive}{13}{black}{0.136}{0.048} & \priorityfullcell{priorityPositive}{80}{white}{0.578}{0.270} & \priorityfullcell{priorityPositive}{41}{black}{0.442}{0.251} \\
\midrule
Granite-4-8B & baseline & \priorityfullcell{priorityPositive}{39}{black}{0.647}{0.090} & \priorityfullcell{priorityPositive}{45}{black}{0.627}{0.095} & \priorityfullcell{priorityPositive}{35}{black}{0.658}{0.229} & \priorityfullcell{priorityNegative}{45}{black}{0.360}{0.092} & \priorityfullcell{priorityNegative}{19}{black}{0.094}{0.060} & \priorityfullcell{priorityNegative}{66}{white}{0.137}{0.091} & \priorityfullcell{priorityNegative}{43}{black}{0.158}{0.094} & \priorityfullcell{priorityNegative}{14}{black}{0.523}{0.242} & \priorityfullcell{priorityPositive}{46}{black}{0.596}{0.145} \\
 & A & \priorityfullcell{priorityNegative}{41}{black}{0.508}{0.059} & \priorityfullcell{priorityNegative}{40}{black}{0.492}{0.054} & \priorityfullcell{priorityNegative}{60}{white}{0.486}{0.430} & \priorityfullcell{priorityPositive}{38}{black}{0.516}{0.058} & \priorityfullcell{priorityNegative}{20}{black}{0.244}{0.194} & \priorityfullna & \priorityfullna & \priorityfullna & \priorityfullna \\
 & P1 & \priorityfullcell{priorityPositive}{26}{black}{0.638}{0.099} & \priorityfullcell{priorityPositive}{28}{black}{0.616}{0.105} & \priorityfullcell{priorityPositive}{24}{black}{0.659}{0.227} & \priorityfullcell{priorityNegative}{27}{black}{0.373}{0.103} & \priorityfullcell{priorityNegative}{18}{black}{0.089}{0.051} & \priorityfullcell{priorityNegative}{45}{black}{0.100}{0.066} & \priorityfullcell{priorityNegative}{26}{black}{0.136}{0.087} & \priorityfullcell{priorityPositive}{51}{black}{0.545}{0.258} & \priorityfullcell{priorityPositive}{63}{white}{0.557}{0.247} \\
 & P2a & \priorityfullcell{priorityNegative}{18}{black}{0.600}{0.103} & \priorityfullcell{priorityNegative}{27}{black}{0.547}{0.140} & \priorityfullcell{priorityNegative}{23}{black}{0.600}{0.249} & \priorityfullcell{priorityPositive}{19}{black}{0.420}{0.114} & \priorityfullcell{priorityPositive}{15}{black}{0.840}{0.197} & \priorityfullcell{priorityNegative}{22}{black}{0.082}{0.074} & \priorityfullcell{priorityNegative}{24}{black}{0.142}{0.106} & \priorityfullcell{priorityNegative}{29}{black}{0.286}{0.243} & \priorityfullcell{priorityPositive}{33}{black}{0.400}{0.280} \\
 & P2b & \priorityfullcell{priorityPositive}{14}{black}{0.641}{0.098} & \priorityfullcell{priorityPositive}{15}{black}{0.618}{0.101} & \priorityfullcell{priorityPositive}{13}{black}{0.671}{0.221} & \priorityfullcell{priorityNegative}{15}{black}{0.371}{0.102} & \priorityfullcell{priorityPositive}{11}{black}{0.186}{0.168} & \priorityfullcell{priorityPositive}{9}{black}{0.098}{0.103} & \priorityfullcell{priorityNegative}{17}{black}{0.095}{0.047} & \priorityfullcell{priorityPositive}{44}{black}{0.472}{0.287} & \priorityfullcell{priorityPositive}{77}{white}{0.549}{0.280} \\
\midrule
Qwen3-8B (ref.) & baseline & \priorityfullcell{gray}{14}{black}{0.556}{0.096} & \priorityfullcell{gray}{14}{black}{0.517}{0.105} & \priorityfullcell{gray}{14}{black}{0.580}{0.220} & \priorityfullcell{gray}{14}{black}{0.469}{0.117} & \priorityfullcell{gray}{14}{black}{0.127}{0.087} & \priorityfullcell{gray}{14}{black}{0.308}{0.217} & \priorityfullcell{gray}{14}{black}{0.261}{0.160} & \priorityfullcell{gray}{14}{black}{0.540}{0.189} & \priorityfullcell{gray}{14}{black}{0.486}{0.178} \\
 & A & \priorityfullcell{gray}{14}{black}{0.604}{0.116} & \priorityfullcell{gray}{14}{black}{0.587}{0.116} & \priorityfullcell{gray}{14}{black}{0.640}{0.210} & \priorityfullcell{gray}{14}{black}{0.427}{0.150} & \priorityfullcell{gray}{14}{black}{0.279}{0.126} & \priorityfullna & \priorityfullna & \priorityfullna & \priorityfullna \\
 & P1 & \priorityfullcell{gray}{14}{black}{0.585}{0.099} & \priorityfullcell{gray}{14}{black}{0.557}{0.099} & \priorityfullcell{gray}{14}{black}{0.611}{0.210} & \priorityfullcell{gray}{14}{black}{0.429}{0.112} & \priorityfullcell{gray}{14}{black}{0.117}{0.088} & \priorityfullcell{gray}{14}{black}{0.209}{0.126} & \priorityfullcell{gray}{14}{black}{0.189}{0.109} & \priorityfullcell{gray}{14}{black}{0.420}{0.178} & \priorityfullcell{gray}{14}{black}{0.395}{0.142} \\
 & P2a & \priorityfullcell{gray}{14}{black}{0.629}{0.100} & \priorityfullcell{gray}{14}{black}{0.603}{0.098} & \priorityfullcell{gray}{14}{black}{0.645}{0.180} & \priorityfullcell{gray}{14}{black}{0.387}{0.111} & \priorityfullcell{gray}{14}{black}{0.820}{0.188} & \priorityfullcell{gray}{14}{black}{0.124}{0.109} & \priorityfullcell{gray}{14}{black}{0.190}{0.100} & \priorityfullcell{gray}{14}{black}{0.347}{0.146} & \priorityfullcell{gray}{14}{black}{0.326}{0.163} \\
 & P2b & \priorityfullcell{gray}{14}{black}{0.623}{0.103} & \priorityfullcell{gray}{14}{black}{0.596}{0.100} & \priorityfullcell{gray}{14}{black}{0.655}{0.199} & \priorityfullcell{gray}{14}{black}{0.391}{0.117} & \priorityfullcell{gray}{14}{black}{0.178}{0.126} & \priorityfullcell{gray}{14}{black}{0.095}{0.103} & \priorityfullcell{gray}{14}{black}{0.120}{0.064} & \priorityfullcell{gray}{14}{black}{0.367}{0.123} & \priorityfullcell{gray}{14}{black}{0.346}{0.200} \\
\bottomrule
\end{tabular}
\caption{Absolute Phase~2 joint-factor results (mean $\pm$ SD).}
\label{tab:phase2_joint_full}
\end{table*}

\begin{table*}[t]
\centering
\small
\setlength{\tabcolsep}{0.5pt}
\renewcommand{\arraystretch}{0.78}
\setlength{\aboverulesep}{0pt}
\setlength{\belowrulesep}{0pt}
\begin{tabular}{l@{\hspace{4pt}}lccccccccc}
\toprule
& & \multicolumn{4}{c}{Performance residuals} & \multicolumn{5}{c}{Position-sensitivity residuals} \\
Family & Method & Acc. & F1 & $\rho$ & MAE & P1 & P2a & P2b & P3a & P3b \\
\midrule
Inference paradigm & Pairwise & \priorityfullcell{priorityNegative}{17}{black}{-8.94}{2.86} & \priorityfullcell{priorityNegative}{18}{black}{-9.70}{3.73} & \priorityfullcell{priorityNegative}{14}{black}{-6.23}{3.57} & \priorityfullcell{priorityPositive}{25}{black}{+17.05}{2.42} & \priorityfullcell{priorityPositive}{36}{black}{+27.83}{12.75} & \priorityfullna & \priorityfullna & \priorityfullna & \priorityfullna \\
& Listwise & \priorityfullcell{priorityNegative}{35}{black}{-27.61}{3.89} & \priorityfullcell{priorityNegative}{43}{black}{-35.10}{4.00} & \priorityfullcell{priorityNegative}{80}{white}{-72.63}{14.13} & \priorityfullcell{priorityPositive}{60}{white}{+52.06}{11.11} & \priorityfullcell{priorityPositive}{10}{black}{+2.11}{31.87} & \priorityfullna & \priorityfullna & \priorityfullna & \priorityfullna \\
& Listwise-compare & \priorityfullcell{priorityNegative}{9}{black}{-0.67}{7.05} & \priorityfullcell{priorityPositive}{9}{black}{+0.55}{5.84} & \priorityfullcell{priorityPositive}{14}{black}{+5.81}{3.79} & \priorityfullcell{priorityPositive}{10}{black}{+2.44}{9.82} & \priorityfullcell{priorityNegative}{26}{black}{-18.11}{6.64} & \priorityfullna & \priorityfullna & \priorityfullna & \priorityfullna \\
\midrule
Aggregation method & Majority vote & \priorityfullcell{priorityNegative}{15}{black}{-7.22}{7.27} & \priorityfullcell{priorityNegative}{13}{black}{-5.46}{7.39} & \priorityfullcell{priorityNegative}{25}{black}{-17.61}{21.19} & \priorityfullcell{priorityPositive}{26}{black}{+18.39}{20.83} & \priorityfullcell{priorityPositive}{44}{black}{+35.89}{12.68} & \priorityfullcell{priorityNegative}{38}{black}{-30.58}{3.96} & \priorityfullcell{priorityNegative}{30}{black}{-22.44}{10.81} & \priorityfullcell{priorityNegative}{75}{white}{-67.43}{5.07} & \priorityfullcell{priorityNegative}{64}{white}{-56.10}{16.97} \\
& Mean score & \priorityfullcell{priorityNegative}{24}{black}{-16.28}{8.46} & \priorityfullcell{priorityNegative}{31}{black}{-23.18}{7.32} & \priorityfullcell{priorityNegative}{27}{black}{-18.84}{9.83} & \priorityfullcell{priorityPositive}{32}{black}{+23.83}{14.27} & \priorityfullcell{priorityNegative}{9}{black}{-1.39}{8.76} & \priorityfullcell{priorityNegative}{52}{black}{-44.69}{12.37} & \priorityfullcell{priorityNegative}{44}{black}{-36.27}{9.72} & \priorityfullcell{priorityNegative}{75}{white}{-67.43}{5.07} & \priorityfullcell{priorityNegative}{64}{white}{-56.10}{16.97} \\
& Weighted sum & \priorityfullcell{priorityNegative}{17}{black}{-8.94}{2.86} & \priorityfullcell{priorityNegative}{18}{black}{-9.71}{3.72} & \priorityfullcell{priorityNegative}{14}{black}{-6.27}{3.53} & \priorityfullcell{priorityPositive}{25}{black}{+17.17}{2.54} & \priorityfullcell{priorityPositive}{35}{black}{+27.22}{12.06} & \priorityfullcell{priorityNegative}{49}{black}{-40.97}{9.15} & \priorityfullcell{priorityNegative}{40}{black}{-32.05}{9.52} & \priorityfullcell{priorityNegative}{75}{white}{-67.43}{5.07} & \priorityfullcell{priorityNegative}{64}{white}{-56.10}{16.97} \\
& Copeland & \priorityfullcell{priorityNegative}{9}{black}{-0.61}{3.57} & \priorityfullcell{priorityPositive}{8}{black}{+0.47}{5.24} & \priorityfullcell{priorityNegative}{10}{black}{-2.51}{2.93} & \priorityfullcell{priorityPositive}{10}{black}{+1.89}{5.04} & \priorityfullcell{priorityPositive}{32}{black}{+24.50}{11.59} & \priorityfullcell{priorityNegative}{49}{black}{-41.25}{8.89} & \priorityfullcell{priorityNegative}{40}{black}{-32.55}{9.41} & \priorityfullcell{priorityNegative}{75}{white}{-67.43}{5.07} & \priorityfullcell{priorityNegative}{64}{white}{-56.10}{16.97} \\
& Bradley--Terry & \priorityfullcell{priorityNegative}{28}{black}{-20.45}{1.10} & \priorityfullcell{priorityNegative}{31}{black}{-23.46}{3.49} & \priorityfullcell{priorityNegative}{50}{black}{-42.12}{8.21} & \priorityfullcell{priorityPositive}{73}{white}{+66.00}{20.62} & \priorityfullcell{priorityPositive}{51}{black}{+43.00}{10.84} & \priorityfullcell{priorityNegative}{22}{black}{-13.97}{5.83} & \priorityfullcell{priorityNegative}{21}{black}{-13.56}{14.01} & \priorityfullcell{priorityNegative}{75}{white}{-67.43}{5.07} & \priorityfullcell{priorityNegative}{64}{white}{-56.10}{16.97} \\
\midrule
Debiasing strategy & Label averaging & \priorityfullcell{priorityNegative}{9}{black}{-1.07}{1.92} & \priorityfullcell{priorityNegative}{10}{black}{-1.63}{1.96} & \priorityfullcell{priorityNegative}{9}{black}{-1.04}{2.33} & \priorityfullcell{priorityPositive}{9}{black}{+0.53}{2.31} & \priorityfullna & \priorityfullna & \priorityfullna & \priorityfullna & \priorityfullna \\
& Demo averaging & \priorityfullcell{priorityNegative}{8}{black}{-0.42}{5.23} & \priorityfullcell{priorityPositive}{8}{black}{+0.20}{6.32} & \priorityfullcell{priorityNegative}{9}{black}{-0.61}{2.64} & \priorityfullcell{priorityPositive}{9}{black}{+0.83}{5.38} & \priorityfullna & \priorityfullna & \priorityfullna & \priorityfullna & \priorityfullna \\
& PriDe & \priorityfullcell{priorityPositive}{9}{black}{+0.90}{1.27} & \priorityfullcell{priorityPositive}{10}{black}{+2.11}{2.10} & \priorityfullcell{priorityPositive}{9}{black}{+1.38}{2.29} & \priorityfullcell{priorityNegative}{9}{black}{-0.60}{1.26} & \priorityfullcell{priorityPositive}{9}{black}{+0.73}{1.48} & \priorityfullna & \priorityfullna & \priorityfullna & \priorityfullna \\
& Context. Calib. & \priorityfullcell{priorityNegative}{11}{black}{-2.87}{4.69} & \priorityfullcell{priorityNegative}{11}{black}{-3.27}{5.74} & \priorityfullcell{priorityNegative}{9}{black}{-1.37}{3.55} & \priorityfullcell{priorityPositive}{13}{black}{+5.13}{7.09} & \priorityfullcell{priorityPositive}{11}{black}{+3.30}{4.59} & \priorityfullna & \priorityfullna & \priorityfullna & \priorityfullna \\
\bottomrule
\end{tabular}
\caption{Full Phase~2 pipeline, pairwise-aggregation, and debiasing residuals (mean $\pm$ SD).}
\label{tab:phase2_pipeline_full}
\end{table*}

%% file: appendix/aggregation_table.tex
\section{Pairwise Aggregation Results}
\label{app:aggregation_results}

Table~\ref{tab:aggregation_residuals} reports the full aggregation-method comparison. We place these detailed results in the appendix so that the main text can focus on the pipeline-level trade-off between predictive performance and label-order stability.

\begin{table*}[t]
\centering
\small
\providecommand{\prioritycell}[5]{}
\providecommand{\priorityna}{}
\renewcommand{\prioritycell}[5]{\colorbox{#1!#2}{\textcolor{#3}{\makebox[3.5em][c]{\shortstack{$#4$\\$(\pm #5)$}}}}}
\renewcommand{\priorityna}{\colorbox{gray!20}{\makebox[3.5em][c]{\shortstack{--\\\phantom{$(\pm 0.000)$}}}}}
\setlength{\tabcolsep}{0.2pt}
\renewcommand{\arraystretch}{1.0}
\begin{tabular}{>{\raggedright\arraybackslash}p{0.10\textwidth}>{\raggedright\arraybackslash}p{0.09\textwidth}ccccccccc}
\toprule
& & \multicolumn{4}{c}{Performance residuals} & \multicolumn{5}{c}{Position-sensitivity residuals} \\
Factor & Level & Acc. & F1 & $\rho$ & MAE & P1 & P2a & P2b & P3a & P3b \\
\midrule
\shortstack[l]{Aggreg.\\method} & Majority vote & \prioritycell{priorityNegative}{15}{black}{-7.22}{7.27} & \prioritycell{priorityNegative}{13}{black}{-5.46}{7.39} & \prioritycell{priorityNegative}{25}{black}{-17.61}{21.19} & \prioritycell{priorityPositive}{26}{black}{+18.39}{20.83} & \prioritycell{priorityPositive}{44}{black}{+35.89}{12.68} & \prioritycell{priorityNegative}{38}{black}{-30.58}{3.96} & \prioritycell{priorityNegative}{30}{black}{-22.44}{10.81} & \prioritycell{priorityNegative}{75}{white}{-67.43}{5.07} & \prioritycell{priorityNegative}{64}{white}{-56.10}{16.97} \\
& Mean score & \prioritycell{priorityNegative}{24}{black}{-16.28}{8.46} & \prioritycell{priorityNegative}{31}{black}{-23.18}{7.32} & \prioritycell{priorityNegative}{27}{black}{-18.84}{9.83} & \prioritycell{priorityPositive}{32}{black}{+23.83}{14.27} & \prioritycell{priorityNegative}{9}{black}{-1.39}{8.76} & \prioritycell{priorityNegative}{52}{black}{-44.69}{12.37} & \prioritycell{priorityNegative}{44}{black}{-36.27}{9.72} & \prioritycell{priorityNegative}{75}{white}{-67.43}{5.07} & \prioritycell{priorityNegative}{64}{white}{-56.10}{16.97} \\
& \shortstack[l]{Weighted\\sum} & \prioritycell{priorityNegative}{17}{black}{-8.94}{2.86} & \prioritycell{priorityNegative}{18}{black}{-9.71}{3.72} & \prioritycell{priorityNegative}{14}{black}{-6.27}{3.53} & \prioritycell{priorityPositive}{25}{black}{+17.17}{2.54} & \prioritycell{priorityPositive}{35}{black}{+27.22}{12.06} & \prioritycell{priorityNegative}{49}{black}{-40.97}{9.15} & \prioritycell{priorityNegative}{40}{black}{-32.05}{9.52} & \prioritycell{priorityNegative}{75}{white}{-67.43}{5.07} & \prioritycell{priorityNegative}{64}{white}{-56.10}{16.97} \\
& Copeland & \prioritycell{priorityNegative}{9}{black}{-0.61}{3.57} & \prioritycell{priorityPositive}{8}{black}{+0.47}{5.24} & \prioritycell{priorityNegative}{10}{black}{-2.51}{2.93} & \prioritycell{priorityPositive}{10}{black}{+1.89}{5.04} & \prioritycell{priorityPositive}{32}{black}{+24.50}{11.59} & \prioritycell{priorityNegative}{49}{black}{-41.25}{8.89} & \prioritycell{priorityNegative}{40}{black}{-32.55}{9.41} & \prioritycell{priorityNegative}{75}{white}{-67.43}{5.07} & \prioritycell{priorityNegative}{64}{white}{-56.10}{16.97} \\
& Bradley--Terry & \prioritycell{priorityNegative}{28}{black}{-20.45}{1.10} & \prioritycell{priorityNegative}{31}{black}{-23.46}{3.49} & \prioritycell{priorityNegative}{50}{black}{-42.12}{8.21} & \prioritycell{priorityPositive}{73}{white}{+66.00}{20.62} & \prioritycell{priorityPositive}{51}{black}{+43.00}{10.84} & \prioritycell{priorityNegative}{22}{black}{-13.97}{5.83} & \prioritycell{priorityNegative}{21}{black}{-13.56}{14.01} & \prioritycell{priorityNegative}{75}{white}{-67.43}{5.07} & \prioritycell{priorityNegative}{64}{white}{-56.10}{16.97} \\
\bottomrule
\end{tabular}
\caption{Pairwise aggregation residuals relative to the baseline.}
\label{tab:aggregation_residuals}
\parbox{0.98\textwidth}{\small\textit{Note.} Values are mean ($\pm$SD) and are multiplied by 100. Red and blue indicate positive and negative residuals, respectively, with darker shading for larger absolute changes. Color indicates direction rather than desirability because lower values are preferable for MAE and PFR.}
\end{table*}